\PassOptionsToPackage{dvipsnames, svgnames, x11names}{xcolor}

\documentclass{article} 
\usepackage{iclr2024_conference,times}
\iclrfinalcopy

\usepackage{amsmath,amsfonts,bm}

\def\eqref#1{equation~\ref{#1}}

\def\1{\bm{1}}

\DeclareMathAlphabet{\mathsfit}{\encodingdefault}{\sfdefault}{m}{sl}
\SetMathAlphabet{\mathsfit}{bold}{\encodingdefault}{\sfdefault}{bx}{n}

\usepackage{url}
\usepackage{amsmath}
\usepackage{amssymb}
\usepackage{graphicx}

\usepackage{bbding}
\makeatletter
  \newcommand\figcaption{\def\@captype{figure}\caption}
  \newcommand\tabcaption{\def\@captype{table}\caption}
\makeatother

\usepackage{booktabs}       
\usepackage{amsfonts}       
\usepackage{nicefrac}       
\usepackage{microtype}      

\definecolor{citecolor}{HTML}{2980b9}
\definecolor{linkcolor}{HTML}{c0392b}
\usepackage[dvipsnames, svgnames, x11names]{xcolor}

\newcommand\blfootnote[1]{%
  \begingroup
  \renewcommand\thefootnote{}\footnote{#1}%
  \addtocounter{footnote}{-1}%
  \endgroup
}

\usepackage{multirow}
\usepackage{makecell}
\usepackage{caption}

\usepackage{adjustbox}
\usepackage{wrapfig}

\usepackage{float}
\usepackage{subfig}

\usepackage{color, colortbl}
\usepackage[hidelinks,breaklinks=true,colorlinks,bookmarks=false,citecolor=citecolor,linkcolor=linkcolor]{hyperref}

\usepackage{wrapfig,lipsum,booktabs}

\usepackage{framed}
\usepackage{makecell}
\usepackage{listings}
\definecolor{lightgray}{rgb}{.9,.9,.9}
\definecolor{darkgray}{rgb}{.4,.4,.4}
\definecolor{purple}{rgb}{0.65, 0.12, 0.82}
\lstdefinelanguage{JavaScript}{
  keywords={break, case, catch, continue, debugger, default, delete, do, else, false, finally, for, function, if, in, instanceof, new, null, return, switch, this, throw, true, try, typeof, var, void, while, with},
  morecomment=[l]{//},
  morecomment=[s]{/*}{*/},
  morestring=[b]',
  morestring=[b]",
  ndkeywords={class, export, boolean, throw, implements, import, this},
  keywordstyle=\color{blue}\bfseries,
  ndkeywordstyle=\color{darkgray}\bfseries,
  identifierstyle=\color{black},
  commentstyle=\color{purple}\ttfamily,
  stringstyle=\color{red}\ttfamily,
  sensitive=true
}
\title{\textcolor{Goldenrod3}{\textit{\textbf{SPHINX:}}} The Joint Mixing of Weights, Tasks, and Visual Embeddings for Multi-modal Large Language Models}

\author{Ziyi Lin$^{1,2*}$, Chris Liu$^{1*}$, Renrui Zhang$^{1,2*}$, Peng Gao$^{1*\dagger\ddagger}$, Longtian Qiu$^{1,3*}$\vspace{0.1cm}\\
{\bf Han Xiao$^1$, Han Qiu$^1$, Chen Lin$^1$, Wenqi Shao$^1$, Keqin Chen$^1$, Jiaming Han$^{1,2}$}\vspace{0.1cm}\\
{\bf Siyuan Huang$^1$, Yichi Zhang$^1$, Xuming He$^3$, Hongsheng Li$^{1,2\dagger}$, Yu Qiao$^{1\dagger}$}\vspace{0.1cm}
\\ \\
$^1$Shanghai AI Laboratory,\quad $^2$MMLab, CUHK,\quad $^3$ShanghaiTech University
}

\newcommand{\sphinx}{\textcolor{Goldenrod3}{\textbf{\textit{SPHINX}}}~}

\newcommand{\sphinxonek}{\textcolor{Goldenrod3}{\textbf{\textit{SPHINX-1k}}}~}
\newcommand{\sphinxtwok}{\textcolor{Goldenrod3}{\textbf{\textit{SPHINX-2k}}}~}
\newcommand{\ie}{{\it i.e.}}

\begin{document}

\maketitle
\blfootnote{$^*$ Equal contribution, $^\dagger$ Equal advisory, $^\ddagger$ Project leader}

\begin{abstract}
We present \textcolor{Goldenrod3}{\textbf{\textit{SPHINX}}}, a versatile multi-modal large language model (MLLM) with a joint mixing of model weights, tuning tasks, and visual embeddings. 
First, for stronger vision-language alignment, we unfreeze the large language model (LLM) during pre-training, and introduce a weight mix strategy between LLMs trained by real-world and synthetic data. By directly integrating the weights from two domains, the mixed LLM can efficiently incorporate diverse semantics with favorable robustness.
Then, to enable multi-purpose capabilities, we mix a variety of tasks for joint visual instruction tuning, and design task-specific instructions to avoid inter-task conflict. In addition to the basic visual question answering, we include more challenging tasks such as region-level understanding, caption grounding, document layout detection, and human pose estimation, contributing to mutual enhancement over different scenarios. 
Additionally, we propose to extract comprehensive visual embeddings from various network architectures, pre-training paradigms, and information granularity, providing language models with more robust image representations.
%
%
Based on our proposed joint mixing, \textcolor{Goldenrod3}{\textbf{\textit{SPHINX}}} exhibits superior multi-modal understanding capabilities on a wide range of applications.
On top of this, we further propose an efficient strategy aiming to better capture fine-grained appearances of high-resolution images. With a mixing of different scales and high-resolution sub-images, \textcolor{Goldenrod3}{\textbf{\textit{SPHINX}}} attains exceptional visual parsing and reasoning performance on existing evaluation benchmarks. 
We hope our work may cast a light on the exploration of joint mixing in future MLLM research. Code is released at \url{https://github.com/Alpha-VLLM/LLaMA2-Accessory}.
\end{abstract}

\section{Introduction}
Since the era of big data, large language models (LLMs) have attained tremendous strides~\citep{OpenAI2023ChatGPT,OpenAI2023GPT4TR,brown2020language,touvron2023llama,zhang2022opt}, showcasing unprecedented application scenarios and generalization capabilities. To further expand their capacity ceiling, visual images are also introduced as inputs to develop powerful multi-modal large language models (MLLMs)~\citep{zhang2023llama,li2023blip,llava,zhu2023minigpt,zhao2023mmicl}. These methods can not only generate well-organized language responses inherited from LLMs, but also unlock the multi-modal understanding capability for a wide range of applications, such as providing detailed image captions, answering visual questions, localizing different objects on the image, etc.

Existing MLLMs explored various strategies to endow LLMs with visual instruction-following capacities.
\textbf{1)} Freezing the LLMs during pre-training, and only learning a projection network for vision-language alignment, e.g., a simple MLP layer of LLaMA-Adapter V2~\citep{gao2023llama} and an attention-based visual abstractor of mPLUG-Owl~\citep{ye2023mplug}.
\textbf{2)} Constructing training data of new tasks to endow MLLMs with new visual understanding abilities, e.g., referential dialogues of Kosmos-2~\citep{peng2023kosmos} and region-level grounding of Shikra~\citep{chen2023shikra}.
\textbf{3)} Employing advanced image encoders for extracting visual embeddings, e.g., the CLIP encoder~\citep{radford2021learning} in LLaVA~\citep{liu2023visual} and the Q-Former~\citep{li2023blip} in MiniGPT-4~\citep{zhu2023minigpt}.
\begin{figure}[http]
    \centering
    \includegraphics[width=1\textwidth]{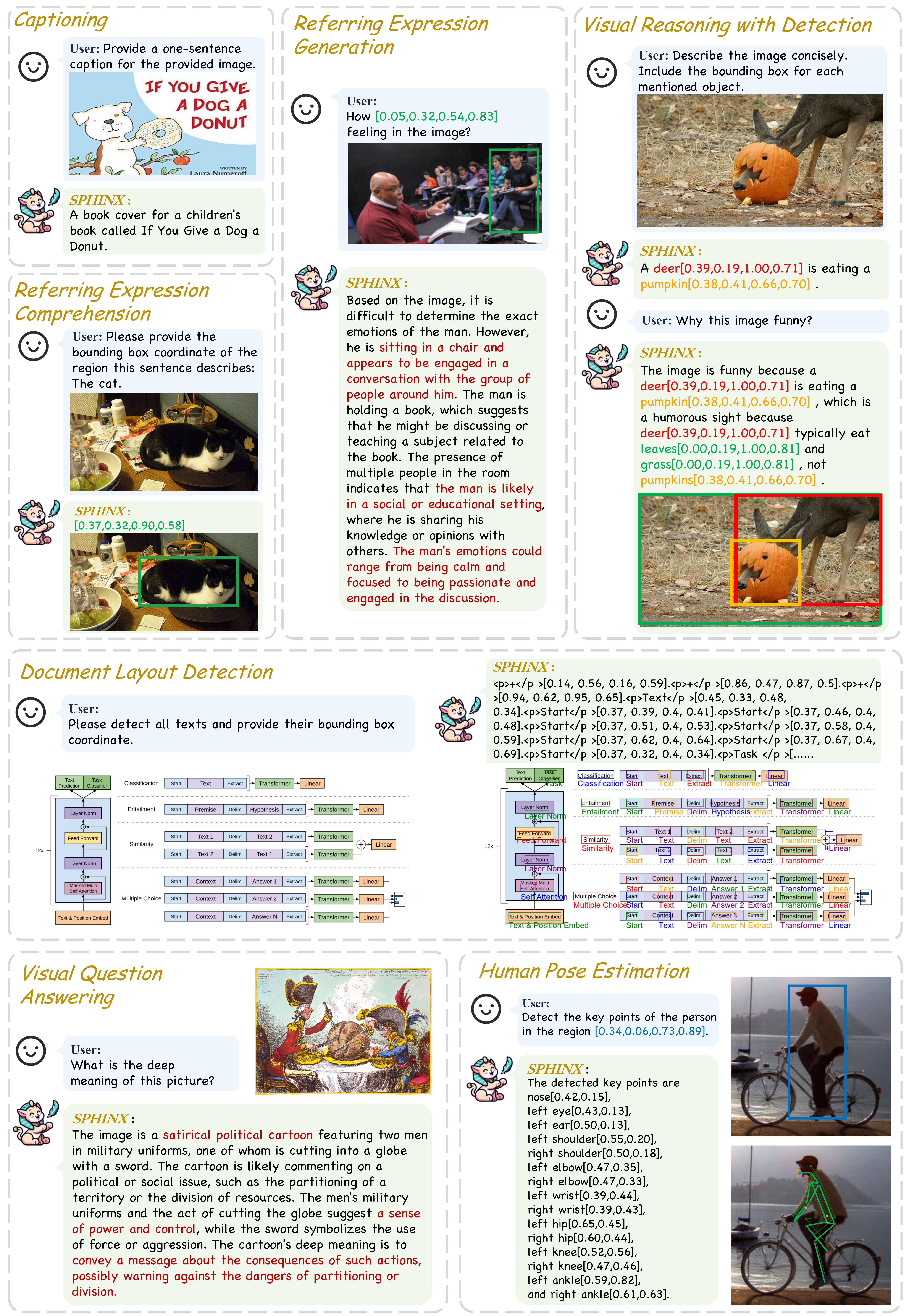}
    \caption{\textbf{Examples of multi-task visual reasoning} by our proposed \textcolor{Goldenrod3}{\textbf{\textit{SPHINX}}}, which excels in diverse visual understanding and perception tasks, such as object detection, caption grounding, and region-level description.}
    \label{fig:demo1}
\end{figure}

\begin{figure}[http]
    \centering
    \includegraphics[width=1\textwidth]{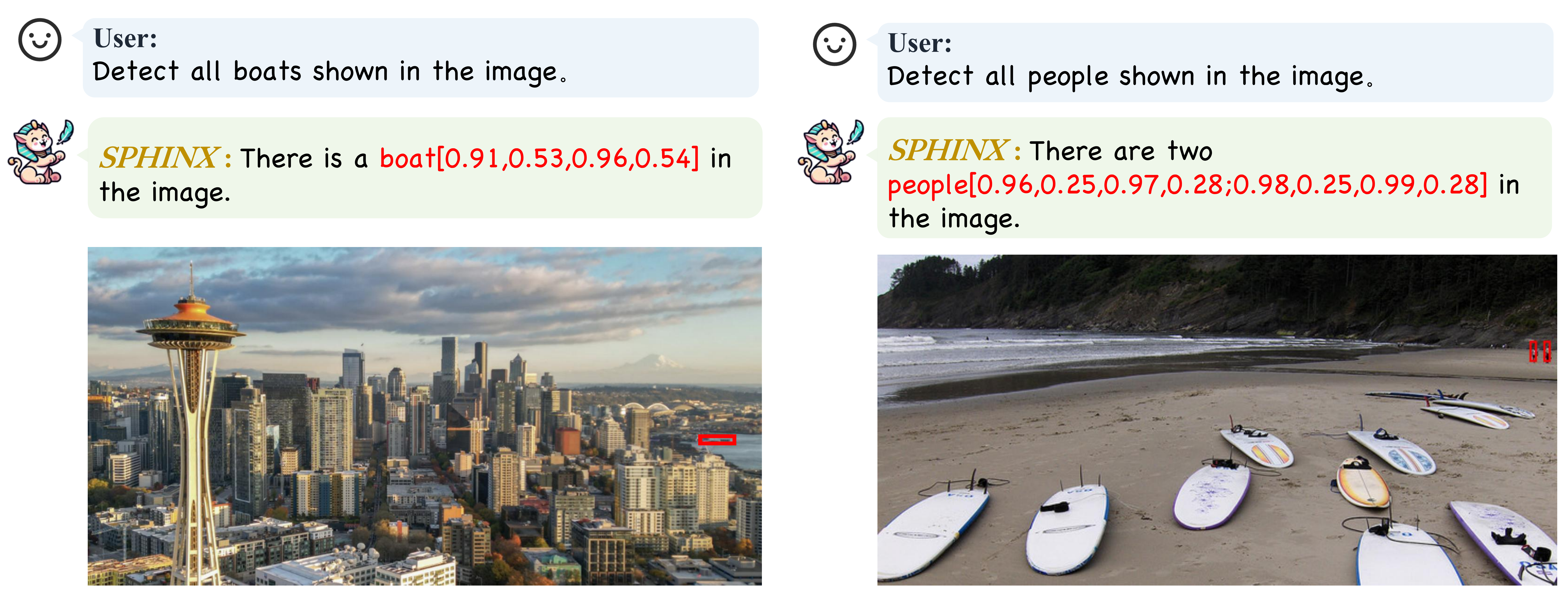}
    \caption{\textbf{Examples of \textcolor{Goldenrod3}{\textbf{\textit{SPHINX}}} for fine-grained visual perception.} With a longer sequence of visual tokens, our model exhibits superior fine-grained understanding capacity.
    }
    \label{fig:demo2}
\end{figure}

In this paper, we propose a versatile MLLM, \textcolor{Goldenrod3}{\textbf{\textit{SPHINX}}}, with a mixing of four significant aspects: model weights, tuning tasks, visual embeddings, and high-resolution sub-images. The main characteristics and findings of our approach is illustrated as follows:

\begin{itemize}
    \item \textbf{Unfreezing LLMs for pre-training.}
    Although the frozen LLM can effectively preserve its long-sentence generation capability, it constrains the potential of better cross-modal alignment via further pre-training on vision-language data. Therefore, we unfreeze the entire LLM, and combine the vision-language datasets~\citep{schuhmann2021laion} for cross-modal alignment and RefinedWeb~\citep{Penedo2023TheRD} for language-specific tuning. This pre-training strategy not only enables LLMs to learn more cross-modal knowledge, but also alleviates the forgetting issue to generate detailed language responses.
    
    \item \textbf{Mixed model weights.}
    Vision-language data from particular domains might contain special semantics, e.g., synthetic captions~\citep{laioncoco} compared to real-world ones~\citep{schuhmann2021laion}. Considering that directly mixing such data might confuse the MLLM, we introduce a weight-mixing strategy to efficiently combine such domain-specific knowledge. Based on the MLLM pre-trained on real-world data, we fine-tune it on the synthetic data, and then linearly combine the finetuned LLM's weights with the real-world ones. In this way, the two types of models would not be affected by contradictory data and our final \textcolor{Goldenrod3}{\textbf{\textit{SPHINX}}} can effectively integrate knowledge from both synthetic and real-world domains.

   \item \textbf{Mixed tuning tasks.} Different from existing task-specific MLLM models~\citep{ye2023mplug,peng2023kosmos,chen2023shikra,llava,gao2023llama}, we integrate a diverse set of visual instruction tasks to tune the pre-trained model, aiming to acquire a wide range of capabilities.
   Our mixing of tasks includes basic visual question answering (VQA), region-level referring expression comprehension/generation (REC/REG), multi-object detection and relation reasoning, text-oriented chart/document VQA, human pose estimation, etc. By such a comprehensive multi-task training paradigm, our \textcolor{Goldenrod3}{\textbf{\textit{SPHINX}}} is a well-performing generalist model for visual instruction following.
   
   \item \textbf{Mixed visual embeddings.} To take the advantage of different encoders, we propose to mix the visual embeddings from various vision backbones~\citep{oquab2023dinov2,li2023blip,radford2021learning} with different network architectures (CNN vs. ViT), pre-training paradigms (supervised vs. self-supervised), and information granularity (global vs. local). By mixing the different image tokens channel-wisely and sequence-wisely, \textcolor{Goldenrod3}{\textbf{\textit{SPHINX}}} obtains stronger visual representations and leads to better vision-language alignment efficacy.

\end{itemize}


On top of this, we further investigate another challenging issue within existing MLLMs, i.e., the limited resolution of input images. As the pre-trained image encoders normally adopt a relatively low image resolution, e.g., 224$\times$224, it severely hinders fine-grained visual comprehension and reasoning for MLLMs. However, simply upsampling the images for encoders would harm the pre-trained positional prior, and, more importantly, lead to expensive computational overhead (the complexity increases quadratically to image size in self-attention mechanisms). 
Therefore, we propose to endow \textcolor{Goldenrod3}{\textbf{\textit{SPHINX}}} with a longer sequence of visual embeddings of mixing different scales and high-resolution sub-images. 
\begin{itemize} 
\item \textbf{Mixed scales and high-resolution sub-images.} we first spatially divide the input high-resolution image into multiple sub-images, and also downsample it into a low-resolution one. Then, we feed all the images concurrently into the mixed visual encoders, and concatenate the extracted multiple token groups to represent the entire high-resolution visual features. 
By mixing visual embeddings of different scales and sub-images, our \textcolor{Goldenrod3}{\textbf{\textit{SPHINX}}} can adaptively explore more fine-grained visual semantics from the high resolution and multi-scale image representations, while maintaining encoding efficiency.
\end{itemize}

Note that, as the different sub-images of high-resolution images do not interact with each other in the visual encoder, they are forced to interchange information within the attention layers of LLMs, which motivates LLMs to process visual conditions more thoroughly and deeply. 
By the proposed three-fold mixer along with a longer visual token sequence, \textcolor{Goldenrod3}{\textbf{\textit{SPHINX}}} fine-tunes LLMs, e.g., LLaMA-2~\citep{Touvron2023Llama2O}, to be a powerful MLLM with superior visual instruction-following capacity. As shown by the examples in Figure~\ref{fig:demo1}, our model excels in a variety of vision tasks, e.g., detecting different objects with remarkable precision and parsing their relations, or accurately interpreting the content within complicated figures. Importantly, as shown in Figure~\ref{fig:demo2}, \textcolor{Goldenrod3}{\textbf{\textit{SPHINX}}} can achieve impressive fine-grained visual perception for high-resolution images, which exhibits \textit{state-of-the-art} performance on extensive evaluation benchmarks, e.g., MMBench~\citep{Liu2023MMBenchIY}, MME~\citep{Fu2023MMEAC}, and POPE~\citep{Li2023EvaluatingOH}.

\section{Related Work}
\paragraph{Large language models (LLMs).}
The field of Natural Language Processing (NLP) has witnessed significant progress over the years, particularly with the advent of LLMs. With Transformer~\citep{Vaswani2017AttentionIA} as the fundamental architecture, LLMs~\citep{OpenAI2023ChatGPT,radford2019language,OpenAI2023GPT4TR} have demonstrated unprecedented performance in modeling intricate language patterns over extensive contexts. Therein, BERT~\citep{Devlin2019BERTPO} showcases the benefits of pre-training on vast text corpora and fine-tuning on specific tasks, setting new standards on various benchmarks.
OpenAI's GPT series~\citep{Radford2018ImprovingLU,radford2019language,OpenAI2023ChatGPT,OpenAI2023GPT4TR}, especially GPT-3~\citep{brown2020language}, harness the power of massive model scaling, with billions and even trillions of parameters. To obtain better instruction following ability, InstructGPT~\citep{Ouyang2022TrainingLM} and ChatGPT~\citep{OpenAI2023ChatGPT} are presented to exhibit exceptional fluency and versatility in open-domain conversation tasks,  ranging from text generation to question answering. Recently, the instruction tuning based on LLaMA~\citep{touvron2023llama} and LLaMA-2~\citep{Touvron2023Llama2O} has gained great popularity as open-source LLMs in the community. Therein, Alpaca~\citep{alpaca} and LLaMA-Adapter~\citep{zhang2023llama} respectively adopt full and parameter-efficient fine-tuning to acquire favorable instruction-following LLMs. Vicuna~\citep{vicuna2023} and GPT-4-LLM~\citep{gpt4llm} further showcase the improvement brought by higher-quality instruction datasets. Other efforts also extend LLMs for match problem solving~\citep{wang2023mathcoder,zhou2023solving}, visual model system~\citep{wu2023visual,yang2023mm}, and open-world recognition~\citep{zhang2023prompt,zhu2022pointclip}. In this paper, we develop our \textcolor{Goldenrod3}{\textbf{\textit{SPHINX}}} based on the superior language understanding of LLaMA-2~\citep{Touvron2023Llama2O} and instruction tuning experience of LLaMA-Adapter series~\citep{zhang2023llama,gao2023llama}, which introduce a three-fold mixer to extend the capability ceiling of instruction-following LLMs for multi-modal input.

\paragraph{Multi-modal large language models (MLLMs).}
In addition to language instruction following, many efforts have been made to inject multi-modal conditions into LLMs for wider application scenarios. As prior attempts, VisualGPT~\citep{Chen2021VisualGPTDA} and BLIP series~\citep{li2023blip,Li2022BLIPBL,Dai2023InstructBLIPTG} indicate the potential of aligning LLMs with visual input for image captioning and question answering. Flamingo~\citep{alayrac2022flamingo} and Kosmos-1~\citep{huang2023kosmos} further exhibit promising multi-modal understanding performance for image-text interleaved contexts. With large-scale pre-training and model sizes, GPT-4~\citep{OpenAI2023GPT4TR} and Bard~\citep{bard} both showcase remarkable proficiency in vision-language understanding and reasoning over diverse multi-modal tasks.
In parallel, a bunch of works have been proposed to align LLaMA with vision modality for advanced visual instruction-following capabilities. LLaVA~\citep{llava} and MiniGPT-4~\citep{zhu2023minigpt} utilize a simple projection layer to connect vision encoders~\citep{li2023blip,radford2021learning} with LLMs. LLaMA-Adapter V2~\citep{gao2023llamaadapter} introduces zero-initialized attention mechanisms for efficient visual instruction tuning, and mPLUG-Owl~\citep{ye2023mplug} adopts delicately designed intermediate networks for cross-modal alignment. For more modality input, ImageBind-LLM~\citep{han2023imagebind} and PandaGPT~\citep{Su2023PandaGPTOM} further incorporate audio and video conditions guided by ImageBind~\citep{Girdhar2023ImageBindOE}. Besides, recent MLLMs are also extended to region-level parsing~\citep{chen2023shikra,peng2023kosmos}, in-context learning~\citep{Li2023MIMICITMI,Li2023OtterAM}, arbitrary image resolutions~\citep{fuyu-8b}, text-to-image generation~\citep{wen2023improving,dong2023dreamllm}, and 3D question answering~\citep{Xu2023PointLLMEL,Guo2023PointBindP,hong20233d}. Different from previous works, our \textcolor{Goldenrod3}{\textbf{\textit{SPHINX}}} aims for image-conditioned MLLM, and proposes a three-fold mixer, i.e., model weights, tuning tasks, and visual embeddings, attaining superior generalization capacity for multi-modal learning.

\section{SPHINX}

In this section, we introduce a versatile MLLM, \textcolor{Goldenrod3}{\textbf{\textit{SPHINX}}}, with the joint mixing of model weights, tuning tasks, visual embeddings, and high-resolution sub-image tokens in Section~\ref{s3.1} and Section~\ref{s3.2}. Finally, in Section~\ref{s3.3}, we introduce several extended applications of \textcolor{Goldenrod3}{\textbf{\textit{SPHINX}}}.

\subsection{The joint mixing of model weights, tuning tasks, and visual embeddings}
\label{s3.1}

\begin{figure*}[t!]
  \centering
\includegraphics[width=\textwidth]{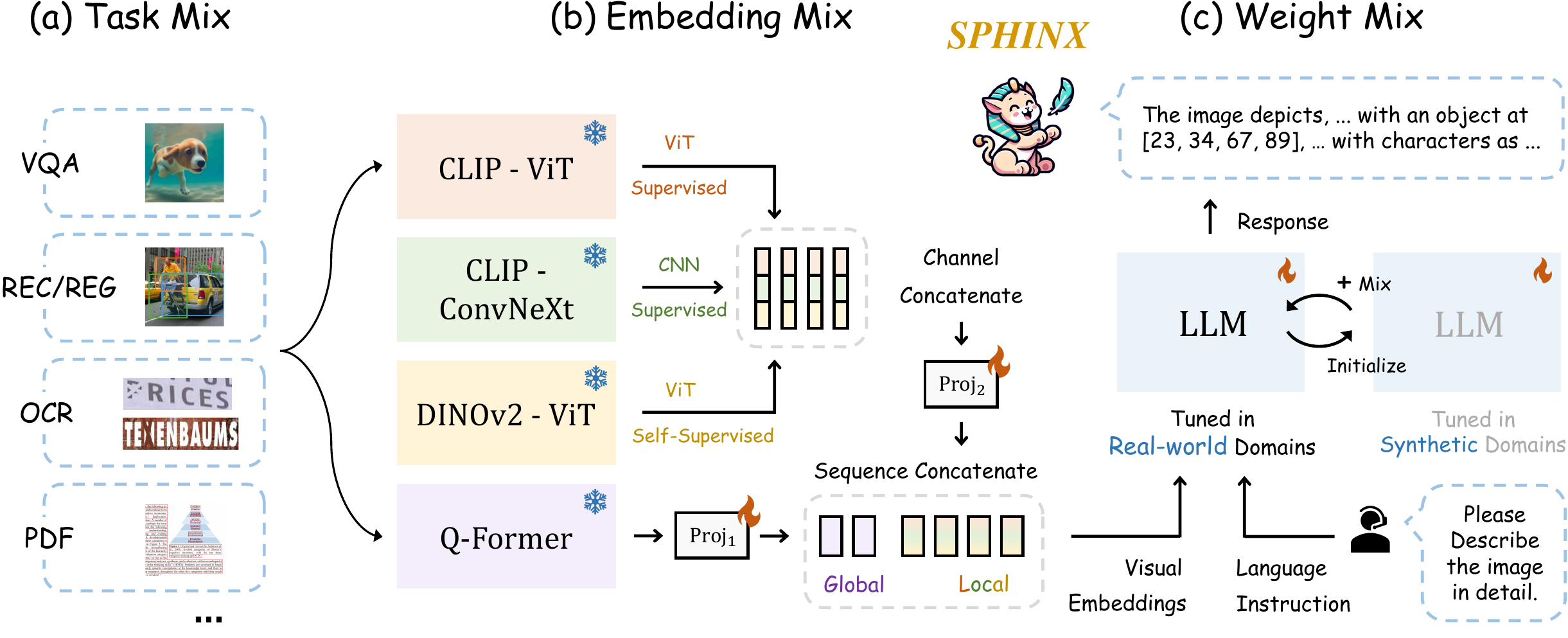}
   \caption{\textbf{The joint mixing paradigm of \textcolor{Goldenrod3}{\textbf{\textit{SPHINX}}}.} with mixed tuning tasks (a), mixed visual embeddings (b), and mixed model weights (c).}
    \label{fig1}
\end{figure*}

The overall mixing paradigm of \textcolor{Goldenrod3}{\textbf{\textit{SPHINX}}} is shown in Figure~\ref{fig1}. We adopt a two-stage training paradigm: the first pre-training stage for vision-language alignment, and the second fine-tuning stage for visual instruction-following learning. During the two stages, we apply the proposed mixing of model weights and tuning tasks, respectively. The model is composed of an LLM, e.g., LLaMA-2~\citep{Touvron2023Llama2O}, a mixing of vision encoders, and two linear projection layers. 

\paragraph{Unfreezing LLM for stage-1 pre-training.}
Existing MLLMs~\citep{zhu2023minigpt,li2023blip,Dai2023InstructBLIPTG} generally freeze the entire LLM during the pre-training by image-caption data, and only train intermediate projection layers for vision-language alignment. This strategy can prevent LLMs from over-fitting to generating only short sentences, since the pre-training caption data mostly contain concise descriptions of images. However, the frozen weights largely constrain the cross-modal learning potential of LLMs with large-scale vision-language data. Therefore, we propose to unfreeze the entire LLM along with learnable linear projection layers, for more sufficient vision-language adaption. On the other hand, the vision encoders are kept frozen for high-quality image representations.
To particularly preserve the long-sentence generation ability of LLM, we supplement the existing pre-training vision-language data with additional text corpora data~\cite{Penedo2023TheRD} for language-only tuning. More specifically, in every iteration, we sample one text and several image-caption data respectively from language and vision-language datasets. 


\paragraph{Mixed model weights of different domains.}
Some vision-language data from particular domains contain distinct semantic knowledge, such as the synthetic captions of LAION-COCO~\citep{laioncoco} compared to real-world descriptions of LAION-400M~\citep{schuhmann2021laion}. We propose a weight mixing strategy of domain-specifically tuned LLMs to integrate respective knowledge from real-world and synthetic data.
We first utilize the most common domain data (LAION-400M~\citep{schuhmann2021laion}) for pre-training, which endows the MLLM with fundamental visual understanding capabilities. Then, we regard such a pre-trained model as the initial checkpoint to further fine-tune the LLM on synthetic domains, e.g., LAION-COCO~\citep{laioncoco}. Finally, to take advantage of the best data domains, we directly conduct a weighted mixing of two LLMs' weights for semantic aggregation.
In detail, we denote the parameters of the fundamental LLM as $\theta_{real}$, and the fine-tuned parameters by synthetic data as $\theta_{syn}$. The mixing process is formulated as
\begin{align}
    \theta_{mix} = \beta\cdot\theta_{real} + (1-\beta)\cdot\theta_{syn},
\end{align}
where $\beta$ denotes the mixing coefficient, and $\theta_{mix}$ represents the mixed LLM weights with aggregated semantics. 
Compared to fusing different domain data for joint pre-training, our weight mix strategy can encourage every MLLM to better learn domain-unique knowledge, and exhibit flexible scalability for any new data domains.

\paragraph{Mixed tuning tasks for stage-2 fine-tuning.} After pre-training and model weight mixing, the MLLM has achieved satisfactory alignment between vision and language data. To further enhance the instruction-following capacity, we collect instruction data from a wide range of multi-modal tasks, and jointly fine-tune the model to learn a vision generalist, instead of a specialist for specific scenarios. 
Previous open-source MLLMs can only perform simple visual question answering (VQA) and single large object referring. In contrast, we enable \textcolor{Goldenrod3}{\textbf{\textit{SPHINX}}} to be jointly fine-tuned with a wide range of tasks, and design a set of task-specific instructions to avoid inter-task conflict. 
The mixed tasks include general VQA, region-level referring expression comprehension/generation (REC/REG), multi-object detection and relation reasoning, text-oriented chart/document VQA, and human pose estimation. 
For example, we adopt ``Detect all objects shown in the image'' for general object detection, and ``Detect all texts and provide their bounding box coordinates'' for document layout detection. Please refer to Table~\ref{table:prompt} for detailed instructions on different benchmarks.
Thanks to the superior reasoning capacity of LLM and proper designs of task prompts, \textcolor{Goldenrod3}{\textbf{\textit{SPHINX}}}, \textit{\textbf{for the first time}}, showcases multi-purpose capabilities of visual understanding and perception, excelling in various application scenarios. 

\paragraph{Mixed embeddings for visual encoding.}
To capture robust visual representations from different aspects, we propose to ensemble a variety of vision backbones for image encoding. The visual backbones with different characteristics are chosen as follows. 
\textbf{1)} Different network architectures. As CNN~\citep{he2016deep} and ViT~\citep{dosovitskiy2020image} mainly aggregate different types of visual appearances, i.e., neighboring dependencies and long-range interactions, we adopt CLIP~\citep{radford2021learning} models respectively with ConvNeXt~\citep{woo2023convnext} and ViT image encoders.
\textbf{2)} Different pre-training paradigms. Supervised training can impose explicit semantic information from textual captions or category labels, while self-supervised learning enforces the model to explore implicit pretext task signals. Thus, we further employ the ViT self-supervised by DINOv2~\citep{oquab2023dinov2} as well as the text-supervised vision encoders, CLIP. 
\textbf{3)} Different information granularity. The aforementioned visual encoders all produce visual tokens in the patch level. To better capture global features, we also adopt Q-Former~\citep{li2023blip} to summarize visual embeddings via querying from the global context. After all the aforementioned encoding, we first channel-wisely concatenate the patch level visual tokens. Then, by using two projection layers for dimension alignment, we spatial-wisely concatenate the representations between those of Q-Former and the other patch-level features. The obtained image tokens are directly placed in front of language instructions, which provide visual context for the language instructions.

\begin{figure*}[t!]
  \centering
\includegraphics[width=0.95\textwidth]{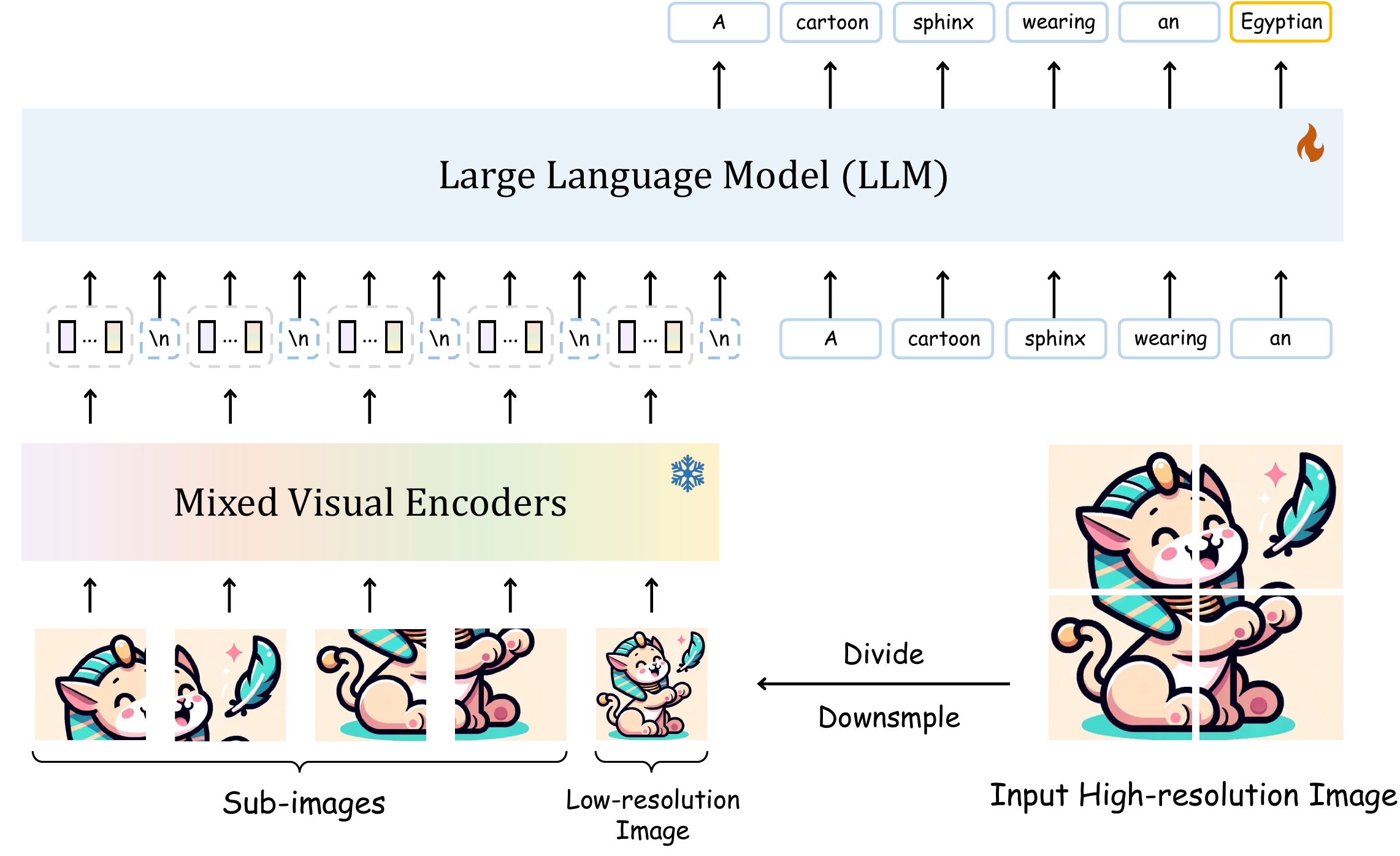}
   \caption{\textbf{Pipeline of \textcolor{Goldenrod3}{\textbf{\textit{SPHINX}}} for high-resolution images.} We propose to further mix different scales and sub-images to better capture fine-grained semantics on high-resolution images.}
    \label{fig2}
\end{figure*}

\subsection{The Mixing of Scales and High-Resolution Sub-images}
\label{s3.2}
With the above-mentioned joint mixing strategy, \textcolor{Goldenrod3}{\textbf{\textit{SPHINX}}} already showcases superior performance for diverse visual perception and reasoning tasks. However, one key challenge remains, i.e., the limited resolution of the input images. To tackle the problem, we further propose to utilize the mixed visual tokens of high-resolution sub-images, as shown in Figure~\ref{fig2}.

\paragraph{Low-resolution constraints of MLLMs.}
State-of-the-art open-source MLLMs~\citep{li2023blip,llava,gao2023llama,chen2023shikra,peng2023kosmos,chen2023minigpt} works adopt frozen image encoders during all training stages, in order to preserve the pre-trained visual semantics. Therefore, the image resolution of MLLMs is usually set as 224$\times$224, severely hindering their efficacy for fine-grained visual perception, especially region-level grounding and description. However, directly processing the upsampled image is not optimal for two reasons. First, to align the image size, the pre-trained positional encoding vectors in ViT are also required to be upsampled correspondingly, which would harm the prior spatial cues. Second, the computation complexity of ViT increases quadratically to the input image size. Thus, naively upsampling the image leads to extensive inference time and GPU memory consumption.

\paragraph{Mixed scales and high-resolution sub-images.}
In our \textcolor{Goldenrod3}{\textbf{\textit{SPHINX}}}, we extend the mixing of visual embeddings to more scales and high-resolution sub-images, allowing for efficient high-resolution image encoding. For an input high-resolution image, e.g., 448$\times$448, we construct five corresponding images of 224$\times$224, and feed them as independent images into our mixed vision encoders. Specifically, we first downsample the input image to 224$\times$224 as an abstract representation, and also downsample the input image to 448$\times$448 and crop four sub-images of 224$\times$224 from the four corners of the 448$\times$448 image, which preserve the detailed visual information. In this way, we enable MLLMs to not only capture fine-grained visual appearances with 224$\times$224 positional encodings, but also achieve favorable computation efficiency. Afterwards, the five groups of image tokens are encoded and concatenated as a long sequence for feeding into LLM, where the first one group encodes global semantics, and the other four record fine-grained local features. 
Importantly, as the image tokens of different patches do not have interaction through the vision encoders, they are forced
to interact within the LLM to obtain complete visual information.
Such a strategy, in turn, motivates LLMs to parse
the relations within visual conditions for better cross-modal learning.
From this perspective, our \textcolor{Goldenrod3}{\textbf{\textit{SPHINX}}} can be regarded as a new paradigm for similar to ViT~\citep{dosovitskiy2020image}, where the mixed vision encoders serve as a patch embedding layer, and the LLM plays the role for patch interaction as a vision decoder.
On visual understanding tasks requiring higher resolutions, \textcolor{Goldenrod3}{\textbf{\textit{SPHINX}}} achieves significant improvement with the mixed visual representations of scales and high-resolution sub-images.

\subsection{Extensions to Wider Applications}
\label{s3.3}
In this section, we respectively introduce some extended applications derived from \textcolor{Goldenrod3}{\textbf{\textit{SPHINX}}}.

\subsubsection{Integration with SAM and Stable Diffusion}

\begin{figure}[t]
    \centering
\includegraphics[width=1\textwidth]{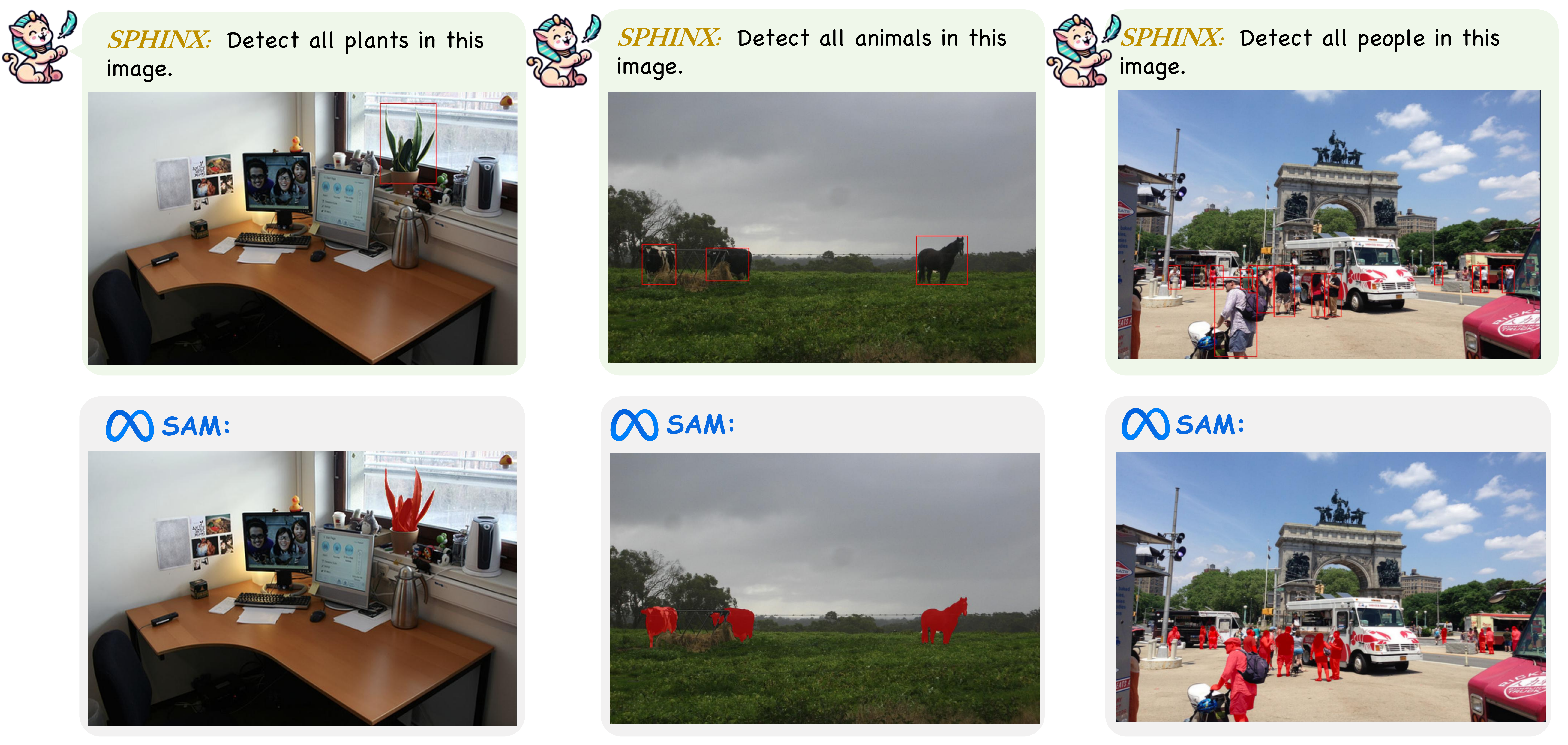}
    \caption{\textbf{Examples of language-referred segmentation} by integrating \textcolor{Goldenrod3}{\textbf{\textit{SPHINX}}} and Segment Anything Model (SAM)~\citep{kirillov2023segment}.
    }
    \label{fig:demo_seg}
\end{figure}

\begin{figure}[t]
    \centering
\includegraphics[width=1\textwidth]{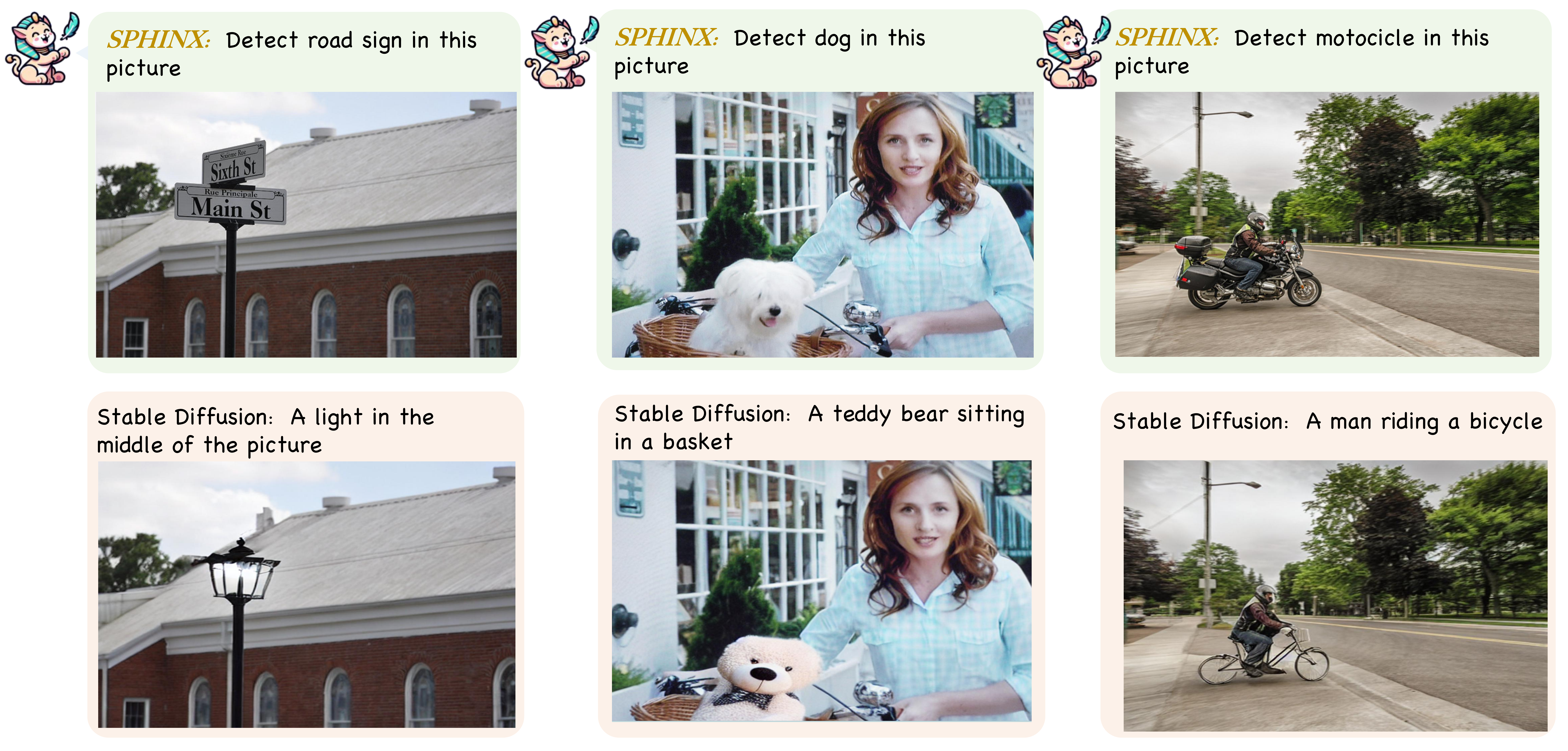}
    \caption{\textbf{Examples of image inpainting and editing} by integrating \textcolor{Goldenrod3}{\textbf{\textit{SPHINX}}} and Stable Diffusion~\citep{rombach2021highresolution}.
    }
    \label{fig:demo_paint}
\end{figure}

In addition to multi-purpose visual instruction-following, we can also integrate \textcolor{Goldenrod3}{\textbf{\textit{SPHINX}}} with other visual foundation models to tackle more challenging vision tasks. Figure~\ref{fig:demo_seg} and~\ref{fig:demo_paint} respectively show two applications for language-referred segmentation and image editing. 

\paragraph{Language-referred segmentation.}
Given that our MLLM is able to output accurate detection boxes with user-provided descriptions or semantic categories, we can cascade the Segment Anything Model (SAM)~\citep{kirillov2023segment} for language-referred instance or semantic segmentation. In detail, we regard the predicted bounding boxes from \textcolor{Goldenrod3}{\textbf{\textit{SPHINX}}} as box prompts, and feed them into SAM for segmenting corresponding instances. In this way, we effectively incorporate the semantic reasoning capability of LLMs and the class-agnostic segmentation of SAM. 

\paragraph{Image inpainting and editing.}
Based on the segmentation results from SAM, we refer to Inpaint Anything~\citep{yu2023inpaint} to integrate image inpainting models (LaMa~\citep{suvorov2021resolution}) and text-to-image generative models (Stable Diffusion~\citep{rombach2021highresolution}) for high-quality image inpainting and editing. Specifically, we first detect and segment the user-indicated objects via \textcolor{Goldenrod3}{\textbf{\textit{SPHINX}}} and SAM as illustrated in the previous paragraph. Then, we feed the segmentation mask into LaMa~\citep{suvorov2021resolution} for removing the corresponding objects with contextual data. After this, the user can prompt Stable Diffusion~\citep{rombach2021highresolution} to further generate new visual content to replace the original ones. This setting integrates our \textcolor{Goldenrod3}{\textbf{\textit{SPHINX}}}, SAM, LaMa, and Stable Diffusion to achieve language-driven image inpainting and editing.

\subsubsection{Fine-tuning SPHINX for Visual Recognition}

Empowered by the joint mixing of weights, tasks and visual embeddings, our \textcolor{Goldenrod3}{\textbf{\textit{SPHINX}}} can comprehend robust and diverse visual category semantics. We propose to regard \textcolor{Goldenrod3}{\textbf{\textit{SPHINX}}} as a universal initialization for traditional visual recognition tasks. For instance, given a classification task of ImageNet-1K~\citep{russakovsky2015imagenet}, we transform the task into a single-turn conversation format of ``Classify the image.'' as the instruction and use ``This is a [CLASS]'' as the response. By performing supervised fine-tuning on the text-converted dataset, we observe fast training convergence on ImageNet-1K. Surprisingly, with only one epoch, \textcolor{Goldenrod3}{\textbf{\textit{SPHINX}}} can achieve 70.8\% classification accuracy without any data augmentation. This convergence speed is much faster than traditional approaches, such as ResNet~\citep{resnet} and ViT~\citep{dosovitskiy2020image} that normally take around 300 training epochs and require strong data augmentation. 


\section{Experiments}

\subsection{Training details}

As mentioned in Section~\ref{s3.1}, our training pipeline consists of two stages. In stage 1, or the Pre-training stage, we start from a text-only LLM, and build the multi-modal capabilities from scratch with large-scale noisy datasets. In stage 2, or the fine-tuning stage, we extract the strong capabilities learned in stage 1 on practical tasks by further training with diverse and high-quality instruct-following datasets. The construct of the datasets and the training configuration for both stages are detailed as follows.

\paragraph{Pre-training datasets.} We use two image captioning datasets LAION-400M~\citep{schuhmann2021laion} and LAION-COCO~\citep{laioncoco} for multi-modal alignment. As we full-fine-tune the language model backbone for long steps, we also jointly train with a text-only dataset RefinedWeb~\citep{Penedo2023TheRD} to avoid harming its text reasoning capability due to catastrophic forgetting.

\paragraph{Pre-training configuration.} We fine-tune the weight of the large language model and the visual projections in the pre-training stage, among which the weight of large language model is initialized from off-the-shelf open-source weights such as LLaMA-2~\citep{Touvron2023Llama2O} and the visual projections are initialized randomly. The visual encoders themselves are kept frozen with their originally pre-trained weights throughout the training. We use the AdamW optimizer~\citep{Kingma2014AdamAM} with $\left( \beta_1, \beta_2 \right) = (0.9, 0.95)$, a cosine annealing learning rate schedule for $180,000$ steps from $5 \times 10^{-5}$ to $5 \times 10^{-6}$ with the first $2,000$ steps being a linear warm-up from $0$ to $5 \times 10^{-5}$, and a constant weight decay of $0.1$. For the joint training on both images and texts, we form each batch with $640$ image-text pairs from LAION-400M or LAION-COCO and $65,536$ text tokens from RefinedWeb. Since captions in LAION-400M and LAION-COCO are based on web-crawled data and generally do not contain much fine-grained information, we only utilize one global view of each image, i.e., the low resolution of 224$\times$224, for faster training. We do not apply any form of language prompts during pre-training. The pre-training time is around 125 hours on 32 A100 GPUs with a 7B language model and about twice the time with a 13B language model.



\paragraph{Fine-tuning datasets.}
In the multi-task fine-tuning phase, our objective is to equip the MLLM with the versatile needs of downstream tasks. Building upon insights from prior research~\citep{llava,Dai2023InstructBLIPTG,chen2023shikra,zhu2023minigpt,Liu2023ImprovedBW}, we include instruction following data such as LLaVA~\citep{llava} and ShareGPT~\citep{ShareGPT}, exposing the model to tasks requiring explicit directives. For general Vision Question Answering (VQA), we leverage datasets like VQAV2~\citep{Agrawal2015VQAVQ} and GQA~\citep{Hudson2019GQAAN}. Expanding the scope to out-of-domain knowledge, we integrate datasets like OKVQA~\citep{Marino2019OKVQAAV} and A-OKVQA~\citep{Schwenk2022AOKVQAAB}, providing the model with information beyond the training data. Optical Character Recognition (OCR) datasets, such as OCRVQA~\citep{Mishra2019OCRVQAVQ} and TextCaps~\citep{Sidorov2020TextCapsAD} are utilized to increase the text understanding ability of \textcolor{Goldenrod3}{\textbf{\textit{SPHINX}}}. We introduce abundant general object detection and pose estimation datasets, such as COCO~\citep{cocodataset} and LVIS~\citep{gupta2019lvis} to inspire the model's capabilities of localization, classification, and human pose estimation. To address grounding tasks, we incorporate RefCOCO~\citep{Kazemzadeh2014ReferItGameRT} and VG~\citep{krishna2017visual} datasets, training the model to handle referring object localization. Additionally, Grounding Caption datasets, such as those from Flickr30k~\citep{Plummer2015Flickr30kEC}, further refine the understanding of descriptions in the context of image regions. Despite the diversity of data sources, we streamline the training by converting all datasets into a multi-turn conversation format. This not only reduces training costs but also enhances overall efficiency. 

\begin{figure}
    \centering
    \includegraphics[width=0.8\textwidth]{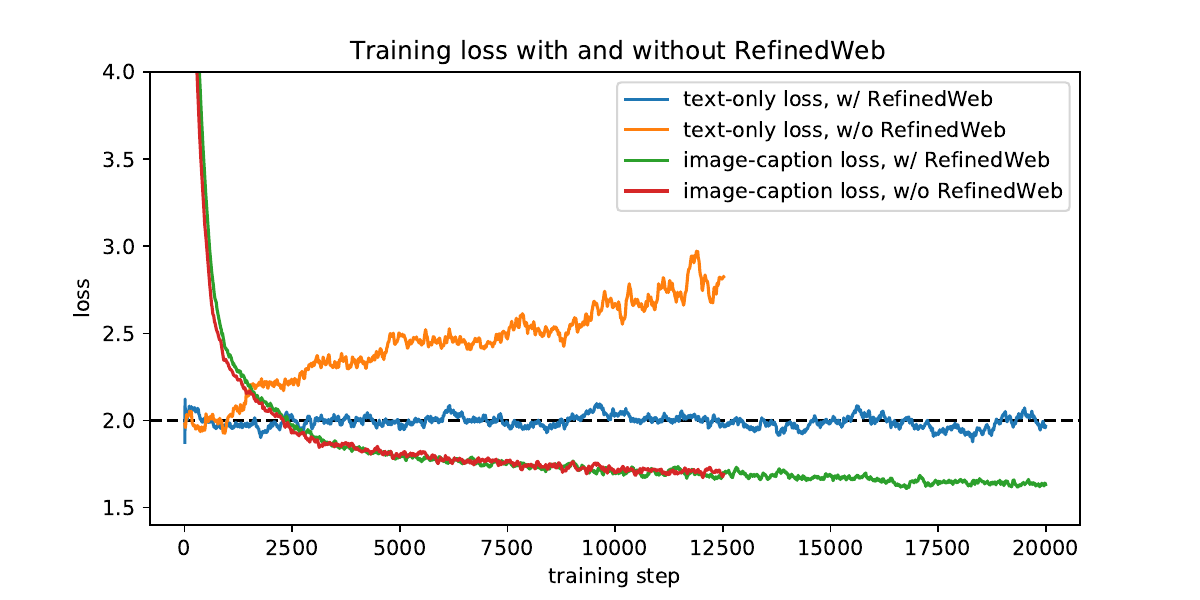}
    \caption{{\bf Loss curve in the pre-training stage with and without optimizing on RefinedWeb.} The text-only loss corresponds to training only on training only RefinedWeb and the image-caption loss corresponds to training only on LAION-400M. Without joint training on RefinedWeb, the image-caption loss descends similarly but the text-only loss grows significantly even in less than 1/10 of the training schedule. We early-stop the {\it without-RefinedWeb} experiments after the forgetting phenomenon is obvious.}
    \label{fig:pretrain_loss_wo_rweb}
\end{figure}

\paragraph{Fine-tuning configuration.}
The trained and frozen network components are identical as the pre-training stage. The optimizer settings are similar to the pre-training stage, except that
we use a batch size of 128, a maximum learning rate of $2 \times 10^{-5}$, a minimum learning rate of 0, and a linear warmup for $0.03$ epoch during fine-tuning.
Training data are sampled from the mixture of datasets following their natural frequencies, {\it i.e.,} the chance of a dataset being sampled from is proportional to its original size.
We follow the image preprocessing steps of~\citep{chen2023shikra,Liu2023ImprovedBW}, {\it i.e.,} padding the image along the shorter edge to make it a square before resizing, for better handling of images with extreme aspect ratios. The fine-tuning takes about 38 hours with 16 A100 GPUs with a 13B language model. The maximum training sequence length is set to 3072.

\subsection{Quantitative evaluation}
In this section, we provide a comprehensive evaluation of \sphinx and showcase results across multiple benchmarks. Our evaluation encompasses both quantitative metrics and qualitative assessments, providing a holistic understanding of our VLM model's performance.

\paragraph{Image-text joint pre-training.} We show in Figure~\ref{fig:pretrain_loss_wo_rweb} the effectiveness of introducing a text-only dataset (\ie, RefinedWeb) to jointly train with image captioning in the pre-training stage. We design an experiment using only vision-language data and without using RefinedWeb. We observe that the text-only loss grows if the model is {\it not} trained with RefinedWeb, showing that our joint-training scheme is effective in preserving the text-modeling capability while adapting for cross-modal understanding.

\paragraph{Evaluation prompt design.}
In our model evaluation, we prioritize aligning with each benchmark's desired output format. To achieve this, we employ distinct prompts tailored to benchmarks that necessitate long answers, short answers, and multiple-choice responses. The detailed information is provided in Table~\ref{table:prompt}. This approach ensures that our model is capable of handling diverse scenarios. 

\begin{table}[]
\adjustbox{max width=\textwidth}{%
\begin{tabular}{c|c}
\toprule
Instructions                                                                                                                                                                   & Benchmarks                               \\ 
\cmidrule(lr){1-1}\cmidrule(lr){2-2}
-                                                                                                                                                                         & LLaVA-Bench, MM-Vet,MathVista            \\ \cmidrule(lr){1-1}\cmidrule(lr){2-2}
Answer the question using a single word or phrase.                                                                                                                        & VQAV2,GQA,OKVQA,VSR,MME,OCR-VQA          \\ \cmidrule(lr){1-1}\cmidrule(lr){2-2}
Answer with the option's letter from the given choices directly.                                                                                                          & SeedBench,ScienceQA,IconVQA              \\ \cmidrule(lr){1-1}\cmidrule(lr){2-2}
Please provide the bounding box coordinate of the region this \\ sentence describes: \{description\}.                          & RefCOCO,RefCOCO+,RefCOCOg                \\ \cmidrule(lr){1-1}\cmidrule(lr){2-2}
Reference OCR token: \{OCR\}\\ Answer the question using a single word or phrase.                                                 & TextVQA                                  \\ \cmidrule(lr){1-1}\cmidrule(lr){2-2}
When the provided information is insufficient, \\ respond with 'Unanswerable'. Answer the question using a \\single word or phrase. & VizWiz                                   \\ \cmidrule(lr){1-1}\cmidrule(lr){2-2}
There are several options: \{options\}                                                                                                                                    & CCBench,MMBench                          \\ \cmidrule(lr){1-1}\cmidrule(lr){2-2}
Detect all objects shown in the image. \\ detect all \{category name\} shown in the image.                                        & Object Detection                         \\ \cmidrule(lr){1-1}\cmidrule(lr){2-2}
Detect all people shown in the image.\\ Detect the key points of the person in the region \{coordinate\}.             & Human Pose Detection \\ \cmidrule(lr){1-1}\cmidrule(lr){2-2}
Detect all texts and provide their bounding box coordinated.  & Document Layout \\ \cmidrule(lr){1-1}\cmidrule(lr){2-2}
Describe the image concisely. \\Include the bounding box for each mentioned object.  & Grounded Caption \\ \cmidrule(lr){1-1}\cmidrule(lr){2-2}
What is the relationship between the object \\in \{coordinate\} and the object in \{coordinate\}?  & Relation Detection \\ \cmidrule(lr){1-1}\cmidrule(lr){2-2}
Please provide the bounding box coordinate of the region this \\ sentence describes: \{description\}  &  Referring Relationship \\ \bottomrule
\end{tabular}
} 
\caption{\textbf{Task-specific instructions on different benchmarks for \sphinx.}}
\label{table:prompt}
\end{table}



\begin{table}[]
\adjustbox{max width=\textwidth}{%
\begin{tabular}{l|cccccccccccc}
\toprule
\multicolumn{1}{c|}{Method} & POPE          & MME\textsuperscript{P}             & MME\textsuperscript{C}             & MMB           & MMB\textsuperscript{CN}           & SEED           & LLava\textsuperscript{W}         & MM-Vet        & CCbench       & MathVista  & Tiny LVLM & Touchstone    \\ 
\cmidrule(lr){1-1}\cmidrule(lr){2-13}
BLIP-2~\citep{li2023blip}                    & 85.3          & 1293.8          & -               & -             & -             & 46.4           & 38.1          & 22.4          & -             & -      & 284.7 &  -      \\
InstructBLIP-7B~\citep{Dai2023InstructBLIPTG}             & -             & -               & -               & 36            & 23.7          & 53.4           & 60.9          & 26.2          & 12.1          & 25.3  & 300.6 & 552.4        \\
InstructBLIP-13B~\citep{Dai2023InstructBLIPTG}            & 78.9          & 1212.8          & -               & -             & -             & -              & 58.2          & 25.6          & -             & -    &  - &  -          \\
Shikra~\citep{chen2023shikra}                      & -             & -               & -               & 58.8          & -             & -              & -             & -             & -             & -    &  - &  -          \\
LLaMA-AdapterV2~\citep{gao2023llamaadapter}             & -             & 1328.40         & 356.43          & -             & -             & -              & -             & -             & -             & -  & 229.2 & 590.1          \\
Qwen-VL-7B~\citep{Bai2023QwenVLAF}                  & -             & -               & -               & 38.2          & 7.4           & 56.3           & -             & -             & 5.5           & -    & -  & -           \\
Qwen-VL-7B-Chat~\citep{Bai2023QwenVLAF}             & -             & 1487.58         & \textbf{360.71} & 60.6          & 56.7          & 58.2           & -             & -             & \textbf{39.3} & -    & \textbf{316.8} & 645.2          \\
LLaVA1.5-7B~\citep{Liu2023ImprovedBW}                 & 85.9          & 1510.7          & -               & 64.3          & 58.3          & 58.6           & 63.4          & 30.5          & 16.4          & -   & - & -            \\
LLaVA1.5-13B~\citep{Liu2023ImprovedBW}                 & 85.9          & 1531.3          & 295.36          & \textbf{67.7} & \textbf{63.6} & 61.6           & 70.7          & 35.4          & 26.5          & -    & - & -           \\
\cmidrule(lr){1-1}\cmidrule(lr){2-13}
\rowcolor[gray]{0.95}
\sphinx                      & 80.7          & 1476.1          & 322.2           & 66.9          & 56.2          & 69.14          & 73.5          & 36.0          & 25.6          & 27.0  & - & 632.4         \\
\rowcolor[gray]{0.95}
\sphinxonek                 & \textbf{90.8} & \textbf{1560.2} & 310.0           & 67.1          & 59.5          & \textbf{71.6}  & 74.3           & 36.6 & 27.9          & 27.5 & 288.9 & 645.0 \\
\rowcolor[gray]{0.95}
\sphinxtwok                 & 87.2           & 1470.6           & 326.8          &   65.9             &   57.9           & \textbf{71.6} & \textbf{76.9}  &   \textbf{40.2}             &     27.4     &  \textbf{27.8} & - & \textbf{659.5} \\
\bottomrule
\end{tabular}
} 
\caption{\textbf{Comparison with SoTA methods on 10 MLLM benchmarks.}}
\label{table:mm}
\end{table}

\paragraph{Model variant definition.} We denote the fundamental variant of our MLLM as \textcolor{Goldenrod3}{\textbf{\textit{SPHINX}}}, which takes as input a low-resolution image of 224$\times$224, and produces 289 visual tokens (257 from the mixed CLIP~\citep{radford2021learning} and DINOv2~\citep{oquab2023dinov2}, and 32 from Q-Former~\citep{li2023blip}). Then, we denote our high-resolution variant as \textcolor{Goldenrod3}{\textbf{\textit{SPHINX-1k}}} and \textcolor{Goldenrod3}{\textbf{\textit{SPHINX-2k}}}. \textcolor{Goldenrod3}{\textbf{\textit{SPHINX-1k}}} processes the image resolution of 448$\times$448 by evenly dividing four sub-images with 1,445 visual tokens, i.e., five groups of 289 tokens (one group for downsampled image and four groups for sub-images). \textcolor{Goldenrod3}{\textbf{\textit{SPHINX-2k}}} further processes a higher resolution of 762$\times$762 with evenly divided nine sub-images of 2,890 visual tokens, i.e., ten groups of 289 tokens.

\paragraph{Benchmarks on multi-modal large language models.}
We test our model on recently proposed MLLM benchmarks to comprehensively evaluation of the model's characteristic such as MME~\citep{fu2023mme}, Seedbench~\citep{Li2023SEEDBenchBM}, POPE~\citep{Li2023EvaluatingOH}, LLaVA-Bench (In-the-Wild)~\citep{llava}, MM-Vet~\citep{Yu2023MMVetEL}, MathVista~\citep{Lu2023MathVistaEM}, MMbench~\citep{liu2023mmbench}, CCbench~\citep{2023opencompass}, Tiny LVLM~\citep{shao2023tiny} and Touchstone~\citep{bai2023touchstone}. We show the result in Table~\ref{table:mm}. We observe that the \sphinx surpasses previous state-of-the-art MLLM performances on 6 out of 10 benchmarks. We compare our model with strong baselines including BLIP-2~\citep{li2023blip}, InstructBLIP~\citep{Dai2023InstructBLIPTG}, Shikra~\citep{chen2023shikra}, Qwen~\citep{Bai2023QwenVLAF}, Fuyu~\citep{fuyu-8b} and LLaVA1.5~\citep{Liu2023ImprovedBW}. The gap between \sphinx and \sphinxonek on POPE suggests that the introduction of high-resolution sub-images can significantly improve visual hallucination problems. 

\paragraph{Visual question answering.}
Furthermore, we evaluate general VQA benchmarks, such as VQAV2~\citep{Agrawal2015VQAVQ}, OKVQA~\citep{Marino2019OKVQAAV}, GQA~\citep{Hudson2019GQAAN}, vizwiz~\citep{Gurari2018VizWizGC}, ScienceQA~\citep{lu2022learn}, visual spatial reasoning (VSR)~\citep{Liu2022VisualSR}, IconQA~\citep{lu2021iconqa}. Additionally, we conduct experiments on Text-oriented VQA such as TextVQA~\citep{Singh2019TowardsVM}, OCR-VQA~\citep{Mishra2019OCRVQAVQ}. We provide the results in Table~\ref{table:vqa}. \sphinx achieves comparative results across all benchmarks. We observe that \sphinxonek and \sphinxtwok significantly outperform \sphinx in VQAv2 datasets and text-oriented VQA that demand fine-grained visual information, showcasing the effectiveness of our visual mixed-up approach for achieving high resolution without relying on a visual encoder trained specifically on high-resolution images. Although the performances of \sphinx on text-oriented VQA surpass strong baselines, such as BLIP-2 and InstructBLIP, it is still below Qwen-VL-7B due to the lack of text-related pre-training data. In the future, we will introduce more text-related pre-training datasets.

\begin{table}[t]
\adjustbox{max width=\textwidth}{%
\begin{tabular}{l|ccccccccc}
\toprule
\multicolumn{1}{c|}{}                         & \multicolumn{7}{c}{General VQA}                                                                                                          & \multicolumn{2}{c}{Text-Oriented VQA}         \\
\multicolumn{1}{c|}{\multirow{-2}{*}{Method}} & OKVQA          & VQAV2         & VizWiz        & GQA           & VSR            & ScienceQA     & IconVQA                                & TextVQA       & OCR-VQA                       \\ 
\cmidrule(lr){1-1}\cmidrule(lr){2-8}\cmidrule(lr){9-10}
BLIP-2~\citep{li2023blip}                                         & 45.9           & -             & 19.6          & 41.0          & 50.9           & -             & 40.6                                   & -             & 40.6                          \\
InstructBLIP~\citep{Dai2023InstructBLIPTG}                                  & -              & -             & 33.4          & 49.5          & 52.1           & -             & 44.8                                   & -             & 44.8                          \\
LLaMA-AdapterV2~\citep{gao2023llamaadapter}                               & 49.6          & 70.7           & 39.8          & 45.1          & -              & -             & -                                      & 37.4          & -                             \\
Shikra~\citep{chen2023shikra}                                        & 47.2           & 77.4          & -             & -             & -              & -             & -                                      & -             & -                             \\
Fuyu-8B~\citep{fuyu-8b}                                       & 60.6           & 74.2            & -           & -          & -           & -             & -                                   & -             & -                          \\
MiniGPT-v2~\citep{chen2023minigpt}                                    & 57.8           & -             & \textbf{53.6} & 60.1          & 62.9           & -             & 51.5                                   & -             & -                             \\
Qwen-VL-7B~\citep{Bai2023QwenVLAF}                                    & 58.6           & 79.5          & 35.2          & 59.3          & 63.8           & 67.1          & -                                      & \textbf{63.8} & \textbf{75.7}                 \\
Qwen-VL-7B-Chat~\citep{Bai2023QwenVLAF}                               & 56.6           & 78.2          & 38.9          & 57.5          & 61.5           & 68.2          & -                                      & 61.5          & 70.5                          \\
LLaVA1.5-7B~\citep{Liu2023ImprovedBW}                                   & -              & 78.5          & 50.0            & 62.0            &     -           & 66.8          & -                                      & 58.2          & -                             \\
LLaVA1.5-13B~\citep{Liu2023ImprovedBW}                                  & -              & 80.0            & \textbf{53.6} & \textbf{63.3} &   -             & \textbf{71.6} & -                                      & 61.3          & -                             \\ 
\cmidrule(lr){1-1}\cmidrule(lr){2-8}\cmidrule(lr){9-10}
\rowcolor[gray]{0.95}
\sphinx                                        & 62.1          & 78.1         & 39.9         & 62.6         & 58.5           &  69.3        & \cellcolor[gray]{0.95}50.4          & 51.63         & \cellcolor[gray]{0.95}66.0 \\
\rowcolor[gray]{0.95}
\sphinxonek                                   & 62.2 & 80.2 & 46.8         & 62.9         & \textbf{65.4} & 69.1         & \cellcolor[gray]{0.95}\textbf{52.7} & 58.78         & \cellcolor[gray]{0.95}70.0 \\
\rowcolor[gray]{0.95}
\sphinxtwok                                   & \textbf{62.6} & \textbf{80.7} & 44.9         & 63.1         & 57.1 & 70.6         & 50.5 & 61.19         & 67.8 \\ \bottomrule
\end{tabular}
} 
\caption{\textbf{Performance comparison on 10 academic task-oriented benchmarks.}}
\label{table:vqa}
\end{table}

\paragraph{Visual grounding.}
Table~\ref{table:ref} evaluates \sphinx on REC benchmarks with RefCOCO~\citep{Kazemzadeh2014ReferItGameRT}, RefCOCO+~\citep{Mao2015GenerationAC}, and RefCOCOg~\citep{Mao2015GenerationAC} datasets. \sphinx outperforms most state-of-the-art models, including specialist model G-DINO-L~\cite{Liu2023GroundingDM} and other visual-language generalist models. Compared to a recent strong baseline Qwen-VL-7B~\citep{Bai2023QwenVLAF}, which also leverages the large language model for visual understanding, our model still achieves better results across all splits by a large margin. Moreover, \sphinxonek and \sphinxtwok enable the use of high-resolution input images, leading to consecutive improvement over \sphinx and narrowing down the gap to the strong specialist model UNINEXT, which adopts a larger input image size. These results demonstrate the competitive capability of \sphinx for visual grounding.

\begin{table}[]
\adjustbox{max width=\textwidth}{%
\begin{tabular}{lcccccccc}
\toprule
\multicolumn{1}{c|}{} & \multicolumn{3}{c}{RefCOCO+} & \multicolumn{3}{c}{RefCOCO} & \multicolumn{2}{c}{RefCOCOg} \\
\multicolumn{1}{c|}{\multirow{-2}{*}{Methods}} & val & test-A & test-B & val & \cellcolor[HTML]{FFFFFF}test-A & test-B & val-u     
 & test-u \\ 
\cmidrule(lr){1-1}\cmidrule(lr){2-4}\cmidrule(lr){5-7}\cmidrule(lr){8-9}{\textit{Specialist models}}\\ 
\cmidrule(lr){1-1}\cmidrule(lr){2-4}\cmidrule(lr){5-7}\cmidrule(lr){8-9}

\multicolumn{1}{l|}{UNINEXT~\citep{Yan2023UniversalIP}} & \multicolumn{1}{l}{85.24} & \multicolumn{1}{l}{89.63} & \multicolumn{1}{l}{79.79} & \multicolumn{1}{l}{92.64} & \multicolumn{1}{l}{94.33} & \multicolumn{1}{l}{91.46}  & \multicolumn{1}{l}{88.73} & \multicolumn{1}{l}{89.37} \\
\multicolumn{1}{l|}{G-DINO-L~\citep{Liu2023GroundingDM}} & \multicolumn{1}{l}{82.75} & \multicolumn{1}{l}{88.95} & \multicolumn{1}{l}{75.92} & 90.56 & 93.19 & 88.24 & \multicolumn{1}{l}{86.13} & \multicolumn{1}{l}{87.02} \\ 

\cmidrule(lr){1-1}\cmidrule(lr){2-4}\cmidrule(lr){5-7}\cmidrule(lr){8-9}{\textit{Generalist models}}\\ 
\cmidrule(lr){1-1}\cmidrule(lr){2-4}\cmidrule(lr){5-7}\cmidrule(lr){8-9}

\multicolumn{1}{l|}{VisionLLM-H~\citep{wang2023visionllm}} & - & - & - & - & 86.70 & - & - & - \\
\multicolumn{1}{l|}{OFA-L~\citep{wang2022ofa}} & 68.29 & 76.00 & 61.75 & 79.96 & 83.67 & 76.39 & 67.57 & 67.58 \\
\multicolumn{1}{l|}{Shikra 7B~\citep{chen2023shikra}} & 81.60 & 87.36 & 72.12 & 87.01 & 90.61 & 80.24 & 82.27 & 82.19 \\
\multicolumn{1}{l|}{Shikra 13B~\citep{chen2023shikra}} & 82.89 & 87.79 & 74.41 & 87.83 & 91.11 & 81.81 & 82.64 & 83.16\\
\multicolumn{1}{l|}{MiniGPT-v2 7B~\citep{chen2023minigpt}} & 79.97 & 85.12 & 74.45 & 88.69 & 91.65 & 85.33 & 84.44 & 84.66 \\
\multicolumn{1}{l|}{MiniGPT-v2 7B-chat~\citep{chen2023minigpt}} & 79.58 & 85.52 & 73.32 & 88.06 & 91.29 & 84.30 & 84.19 & 84.31 \\
\multicolumn{1}{l|}{Qwen-VL-7B~\citep{Bai2023QwenVLAF}} & 83.12 & 88.25 & 77.21 & 89.36 & 92.26 & 85.34 & 85.58 & 85.48 \\
\multicolumn{1}{l|}{Qwen-VL-7B-Chat~\citep{Bai2023QwenVLAF}} & 82.82 & 88.59 & 76.79 & 88.55 & 92.27 & 84.51 & 85.96 & 86.32 \\
\cmidrule(lr){1-1}\cmidrule(lr){2-4}\cmidrule(lr){5-7}\cmidrule(lr){8-9}
\rowcolor[gray]{0.95}
\multicolumn{1}{l|}{\cellcolor[gray]{0.95}\sphinx} & 82.77 & 87.29 & 76.85 & 89.15 & 91.37 & 85.13 & 84.87 & 83.65 \\
\rowcolor[gray]{0.95}
\multicolumn{1}{l|}{\cellcolor[gray]{0.95}\sphinxonek} & \textbf{86.64} & \textbf{91.08} & 80.35 & 91.05 & 92.65 & 86.56 & \textbf{88.19} & 88.35 \\ 
\rowcolor[gray]{0.95}\multicolumn{1}{l|}{\cellcolor[gray]{0.95}\sphinxtwok} & 85.51 & 90.62 & \textbf{80.45} & \textbf{91.10} & \textbf{92.88} & \textbf{87.07} & 88.07 & \textbf{88.65} \\
\bottomrule
\end{tabular}
} 
\caption{Performance comparisons~(Top-1 Accuracy@0.5) on the referring expression comprehension task. The best results among generalist models are marked in bold.}
\label{table:ref}
\end{table}

\begin{figure}[t]
    \centering
\includegraphics[width=1\textwidth]{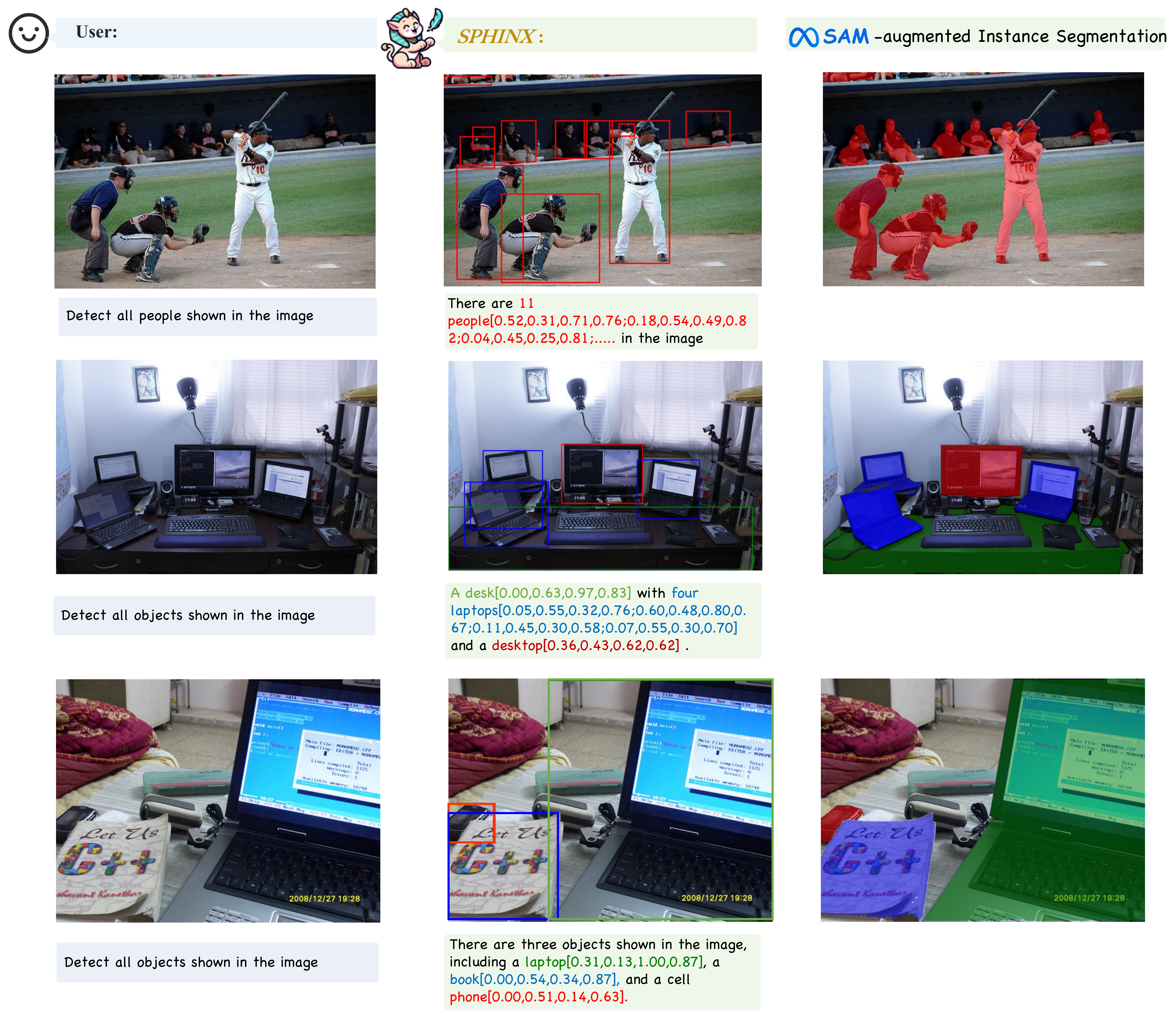}
    \caption{Examples of \textcolor{Goldenrod3}{\textbf{\textit{SPHINX}}} integrating with Segment Anything Model (SAM)~\citep{kirillov2023segment} for language-referred segmentation.}
    \label{fig:demo_SAM1}
\end{figure}

\begin{figure}[http]
    \centering
    \includegraphics[width=1\textwidth]{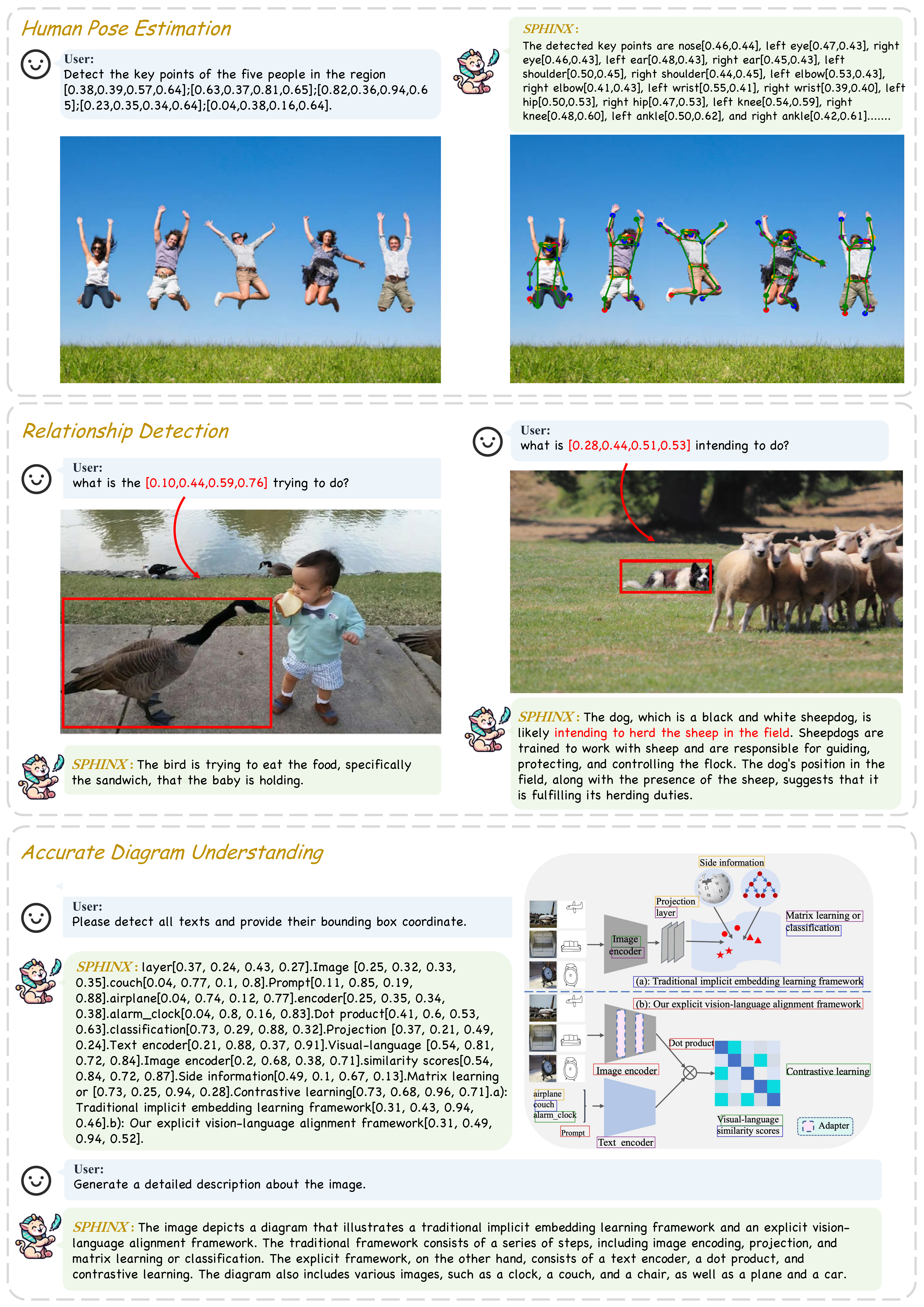}
    \caption{Examples of \textcolor{Goldenrod3}{\textbf{\textit{SPHINX}}} for human pose estimation (Top), relationship reasoning (Middle), and diagram understanding (Bottom).
    }
    \label{fig:demo_2}
\end{figure}
\begin{figure}[h]
    \centering
    \includegraphics[width=1\textwidth]{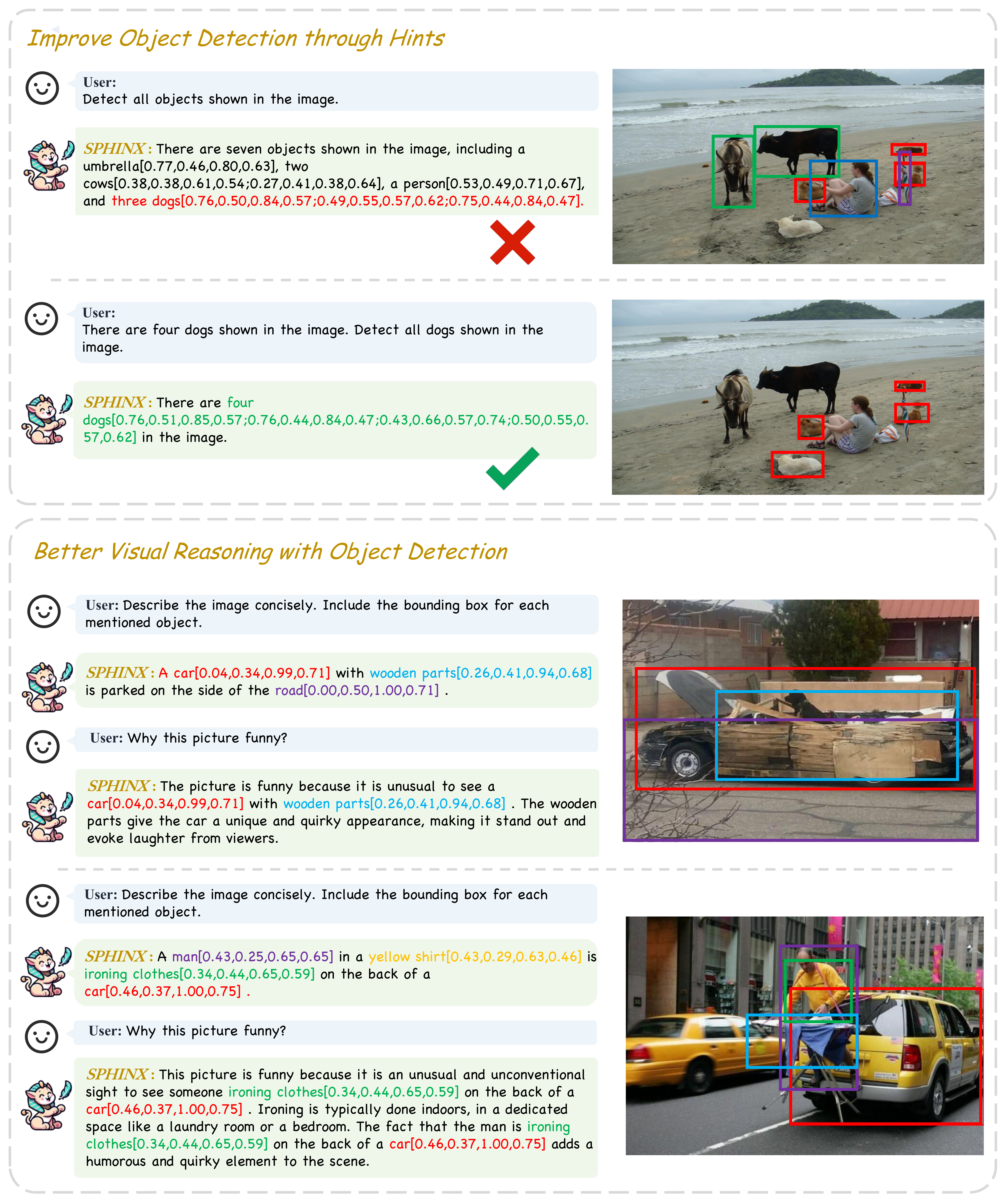}
    \caption{Examples of \textcolor{Goldenrod3}{\textbf{\textit{SPHINX}}} for improved object detection through hints (Top) and better visual reasoning with object detection (Bottom).}
    \label{fig:demo_3}
\end{figure}

\subsection{Demonstrations}
In this section, we present the qualitative outcomes of \textcolor{Goldenrod3}{\textbf{\textit{SPHINX}}}, showcasing its capabilities in SAM-assisted segmentation, general object detection, human pose estimation, document layout detection, anomaly detection, and etc. 
Surprisingly, \sphinx also exhibits improved performance on the chain of thoughts and obtains emergent cross-task abilities.


\paragraph{SAM-augmented instance segmentation.}
We integrate \textcolor{Goldenrod3}{\textbf{\textit{SPHINX}}} with SAM to enhance segmentation capabilities. This integration involves detecting bounding boxes for the target objects and subsequently providing the bounding box coordinates to SAM for the generation of segmentation masks. The results, depicted in Figure \ref{fig:demo_SAM1}, showcase a notable performance improvement achieved through the collaboration of \textcolor{Goldenrod3}{\textbf{\textit{SPHINX}}} and SAM. 
Surprisingly, We observe that the predicted masks for small objects are extremely accurate such as the cell phone in the last row.
The synergistic application of \textcolor{Goldenrod3}{\textbf{\textit{SPHINX}}} and SAM underscores the considerable potential inherent in our methodology.

\paragraph{Region-level understanding.}
In Figure \ref{fig:demo_2}, the performance of \sphinx's detection capabilities is showcased. The upper row displays the synchronized jumping of five teenagers, each assuming distinct poses. Notably, \textcolor{Goldenrod3}{\textbf{\textit{SPHINX}}} accurately predicts the pose with key points for each individual, leaving no participant overlooked. 
The middle row illustrates the \sphinx's reasoning ability to focus on a specified region. We observe that \sphinx successfully recognize the desired objects and detailed answer to the question. 
The bottom row indicates \sphinx's superior diagram understanding ability, which produces accurate layout detection and content comprehension.




\paragraph{Better visual reasoning with object detection.}
The enhanced visual reasoning capabilities of our model with object detection are showcased in Figure \ref{fig:demo_3}. Notably, \sphinx leverages the object detection feedback by initially instructing \sphinx to generate object detection results and then requesting it to answer questions based on localization outcomes. The model will prioritize selecting the most relevant objects for coordinate feedback based on the query content, rather than all detected objects. This underscores the idea that in multi-task training, the synergy between different tasks can significantly enhance overall performance. Furthermore, the model exhibits commendable Contextual Understanding (COT) by effectively integrating information from diverse elements in the image, resulting in more powerful reasoning ability.


\paragraph{Improved object detection through hints.}
We highlight \textcolor{Goldenrod3}{\textbf{\textit{SPHINX}}}'s proficiency in understanding user hints. As depicted in Figure \ref{fig:demo_3}, initially requesting the model to predict all dogs in the image leads to the misidentification of other objects. However, upon offering additional hints about the desired object, \textcolor{Goldenrod3}{\textbf{\textit{SPHINX}}} demonstrates an improved comprehension of instructions and accurately predicts all dogs in the image.

\begin{figure}[t]
    \centering
\includegraphics[width=1\textwidth]{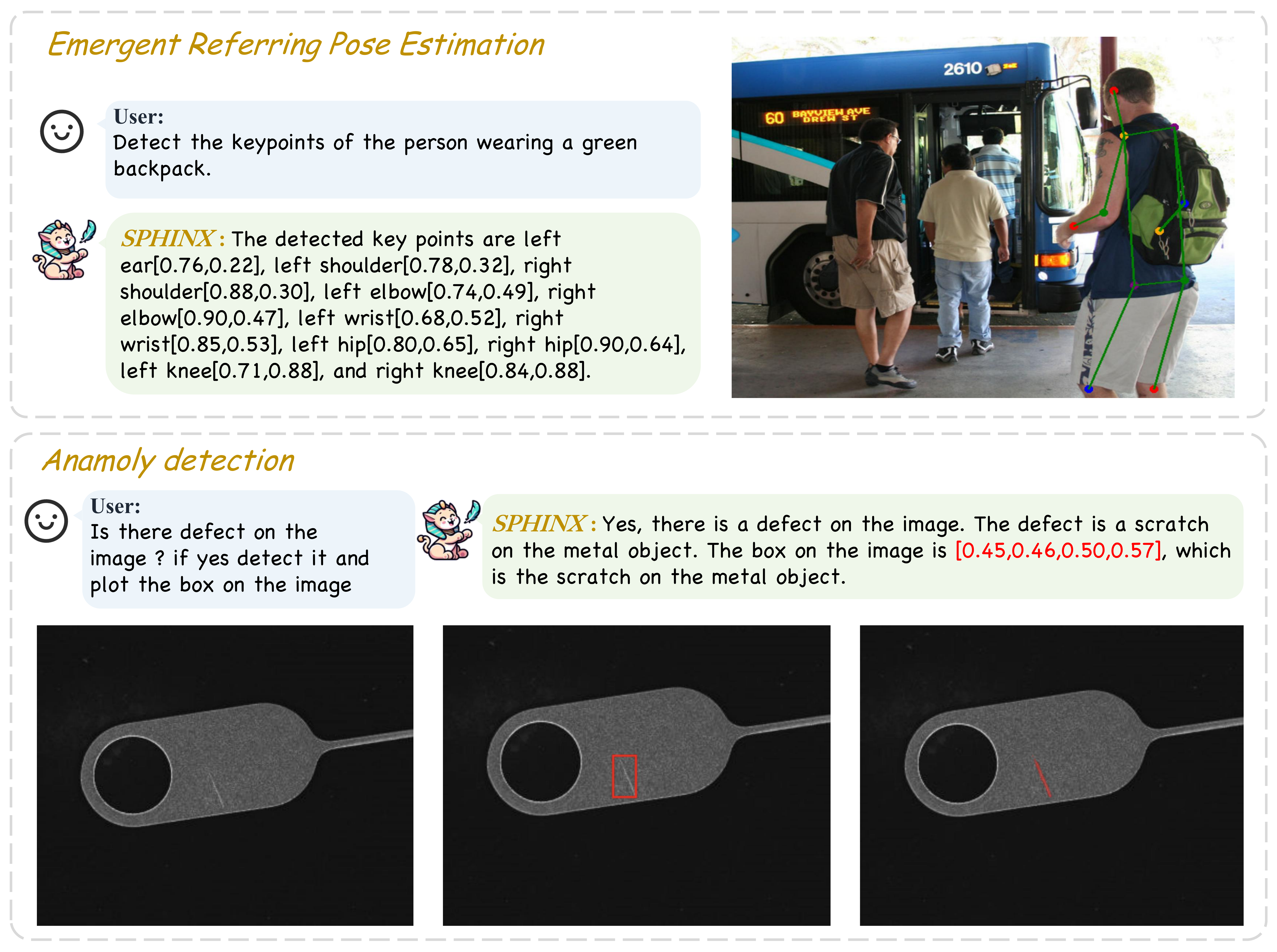}
    \caption{Examples of \textcolor{Goldenrod3}{\textbf{\textit{SPHINX}}} for emergent referring pose estimation (Top) and the challenging anomaly detection (Bottom). 
    }
    \label{fig:demo_refer_pose}
\end{figure}

\paragraph{Emergent referring pose estimation.}
The original referring object comprehension and pose estimation are two different tasks, where the former detects object bounding boxes according to textual descriptions, and the latter outputs human keypoints from given bounding boxes. Interestingly, as shown in Figure~\ref{fig:demo_refer_pose} (Top), by our mixing of the two tuning tasks, our \textcolor{Goldenrod3}{\textbf{\textit{SPHINX}}} acquires the emergent capacity for referring pose estimation, i.e., generating human keypoints directly from textual descriptions. Such an observation indicates that our \textcolor{Goldenrod3}{\textbf{\textit{SPHINX}}} fully comprehend the semantics across different vision-language tasks, and implicitly connect them via superior reasoning power.

\paragraph{Anomaly detection.}
It is important for industrial monitoring and healthcare to detect rare events or outliers that may indicate abnormal or suspicious behavior.
As shown in Figure~\ref{fig:demo_refer_pose} (Bottom), our \textcolor{Goldenrod3}{\textbf{\textit{SPHINX}}} also excels in anomaly detection. Although we do not explicitly involve related training data, our MLLM still demonstrates superior localization accuracy for unsharp defects. This indicates wide potentials of \textcolor{Goldenrod3}{\textbf{\textit{SPHINX}}} in real-world applications.

\begin{figure}[t]
    \centering
\includegraphics[width=1\textwidth]{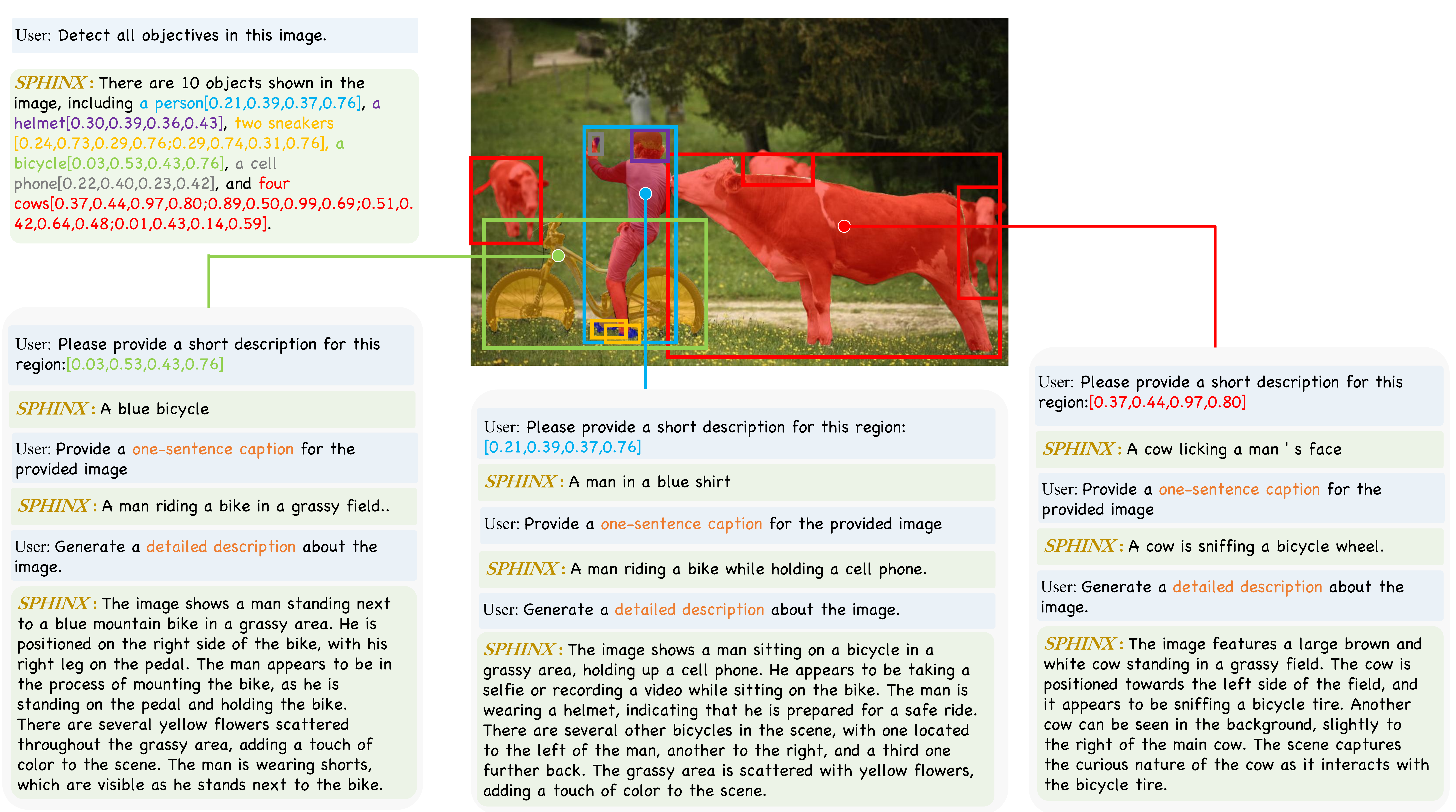}
    \caption{Examples of multi-level dense captioning by \textcolor{Goldenrod3}{\textbf{\textit{SPHINX}}.
    }}
    \label{fig:demo_dense}
\end{figure}

\paragraph{Multi-level dense captioning.}
Endowed with diverse multi-task pre-training, \textcolor{Goldenrod3}{\textbf{\textit{SPHINX}}} can perform multi-level dense captioning by iterative promoting itself. Given an input image, prompting \textcolor{Goldenrod3}{\textbf{\textit{SPHINX}}} with ``Detect all objects shown in the image'' can localize the position of all objects. Then, we iteratively prompt each detected region with ``Please provide a short description for this region : [x1, y1, x2, y2]'' to extract a simple property on the localized region. To get a deeper understanding on the detected regions, we crop all images based on the detection results. Each cropped view is fed independently into \textcolor{Goldenrod3}{\textbf{\textit{SPHINX}}} with two prompts, namely, ``Provide a one-sentence caption for the provided image.'' and ``Generate a detailed description about the image.''. By doing so, we can detect all objects shown in the image and densely label all boxes with property, simple caption, and detailed caption. The multi-level dense captioning results are illustrated in Figure \ref{fig:demo_dense}.

\section{Conclusion}

In this paper, we propose \textcolor{Goldenrod3}{\textbf{\textit{SPHINX}}}, a versatile multi-modal large language model (MLLM) with multi-purpose visual instruction-following capabilities. In our MLLM, we introduce a joint mixing of three different aspects: model weights of pre-trained LLMs by real-world and synthetic data, tuning tasks for diverse visual perception and reasoning tasks, and visual embeddings from different types of vision backbones. On top of this, we further devise to endow \textcolor{Goldenrod3}{\textbf{\textit{SPHINX}}} with the capacity to process high-resolution images by mixing different visual scales and sub-images, which exhibits superior fine-grained visual understanding performance.
Via our proposed three-fold mixing strategy, \textcolor{Goldenrod3}{\textbf{\textit{SPHINX}}} achieves impressive performance over a wide range of multi-modality evaluation benchmarks, and can serve as a strong vision generalist to tackle object detection, region-level captioning, and human pose estimation, etc. Our MLLM can also be integrated with other visual foundation models for wider functionalities, e.g., SAM~\citep{kirillov2023segment} for language-referred segmentation and Stable Diffusion~\citep{rombach2021highresolution} for image editing. Our future work will focus on incorporating a wider range of vision-language tasks into \textcolor{Goldenrod3}{\textbf{\textit{SPHINX}}} for all-purpose capabilities.


		

\bibliography{iclr2024_conference}
\bibliographystyle{iclr2024_conference}

\end{document}